\documentclass[]{worldbench}
\definecolor{w_blue}{RGB}{52,204,204}
\definecolor{w_yellow}{RGB}{255,192,0}

\usepackage{amsfonts}
\usepackage{amsmath}
\usepackage{amssymb}

\usepackage{colortbl}

\usepackage{makecell}
\usepackage{multicol}
\usepackage{multirow}
\usepackage{pifont}
\usepackage[edges]{forest}
\usepackage{wrapfig}

\definecolor{hidden-draw}{RGB}{20,68,106}
\definecolor{hidden-pink}{RGB}{255,245,247}
\usepackage{xspace}

\newcommand*{\eg}{\emph{e.g.}\@\xspace}

\newcommand{\onedot}{.\xspace}
\def\etal{\emph{et al}\onedot}

\definecolor{mygray}{gray}{.9}
\definecolor{mygray1}{gray}{.92}
\definecolor{evaunit01green}{RGB}{82,208,83}
\definecolor{lowred}{RGB}{238,18,137}

\definecolor{lowerred}{RGB}{255,110,180}

\newcommand{\dplus}[1]{\fontsize{6pt}{0.1em}\selectfont (\textbf{\textcolor{lowred}{#1}})}

\definecolor{cameragreen}{RGB}{113,175,71}
\definecolor{lidarblue}{RGB}{103,166,223}
\definecolor{fusionorange}{RGB}{233, 113, 50}

\definecolor{defaultcolor}{RGB}{12,127,17}

\usepackage{pifont}
\newcommand{\cmark}{\ding{51}}
\newcommand{\xmark}{\ding{55}}

\definecolor{vla_purple}{RGB}{138,83,192}
\definecolor{vla_green}{RGB}{78,167,46}

\definecolor{tpami_blue}{RGB}{52,204,204}
\definecolor{tpami_gray}{RGB}{165,165,165}
\definecolor{tpami_red}{RGB}{192,0,0}
\definecolor{tpami_yellow}{RGB}{248,190,50}
\definecolor{link}{RGB}{248,190,50}

\usepackage{titletoc}

\definecolor{crArgoverse}{RGB}{220,100,180}

\definecolor{crBDD100}{RGB}{150,170,220}
\definecolor{crBDDX}{RGB}{110,180,200}
\definecolor{crBenchDrive}{RGB}{210,170,115}
\definecolor{crDriveBench}{RGB}{130,140,90}

\definecolor{crCarla}{RGB}{230,170,120}
\definecolor{crCarlaSC}{RGB}{150,140,110}
\definecolor{crCoVLA}{RGB}{240,210,100}
\definecolor{crCamOcc}{RGB}{220,140,120}
\definecolor{crOpenDV}{RGB}{235,140,120}

\definecolor{crDriveMLM}{RGB}{210,160,170}
\definecolor{crDriveLM}{RGB}{110,110,130}
\definecolor{crDriveCoT}{RGB}{180,140,120}
\definecolor{crDriveAction}{RGB}{140,150,110}

\definecolor{crHBD}{RGB}{200,220,80}

\definecolor{crImpromptuVLA}{RGB}{245,170,190}

\definecolor{crnuScenes}{RGB}{100,170,150}
\definecolor{crnuPlan}{RGB}{170,200,120}
\definecolor{crNoc}{RGB}{180,170,140}
\definecolor{crNAVSIM}{RGB}{140,160,180}
\definecolor{crNuInstruct}{RGB}{220,120,140}
\definecolor{crNuInteract}{RGB}{190,90,130}

\definecolor{crOpenOcc}{RGB}{170,140,190}
\definecolor{crOccThreeD}{RGB}{220,130,110}
\definecolor{crOmniDrive}{RGB}{130,180,130}
\definecolor{crOmniReasonN}{RGB}{100,150,110}
\definecolor{crOmniReasonB}{RGB}{160,190,220}

\definecolor{crPrivate}{RGB}{100,100,100}
\definecolor{crProcGen}{RGB}{90,130,130}
\definecolor{crPhysicalAI}{RGB}{100,120,32}

\definecolor{crRoboBEV}{RGB}{200,110,80}
\definecolor{crReasonDrive}{RGB}{100,150,110}

\definecolor{crSDN}{RGB}{90,140,180}
\definecolor{crSUPAD}{RGB}{190,140,210}

\definecolor{crTalkCar}{RGB}{140,90,170}

\definecolor{crVLAAD}{RGB}{240,180,80}

\definecolor{crWaymo}{RGB}{135,180,225}
\definecolor{crWOMDR}{RGB}{220,100,100}
\definecolor{crWODEE}{RGB}{100,220,150}

\definecolor{crLMDrive}{RGB}{160,180,100}
\definecolor{crLyft}{RGB}{110,160,120}

\definecolor{crMetaAD}{RGB}{190,120,120}

\usepackage{tabularx} 
\usepackage{enumitem}

\title{
Forging Spatial Intelligence: A Roadmap of Multi-Modal Data Pre-Training for Autonomous Systems
}

\author[]{Song~Wang$^{1,}$\raisebox{0.2em}{\includegraphics[width=0.02\linewidth]{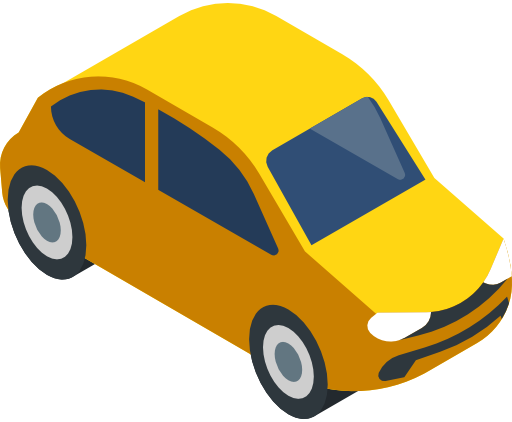}}}
\author[]{Lingdong~Kong$^{2,}$\raisebox{0.2em}{\includegraphics[width=0.02\linewidth]{figs/icons/car1.png}}}
\author[]{Xiaolu~Liu$^{1,}$\raisebox{0.2em}{\includegraphics[width=0.02\linewidth]{figs/icons/car1.png}}}
\author[]{Hao~Shi$^{1}$}
\author[]{Wentong~Li$^{3}$}
\author[]{Jianke~Zhu$^{1,}$\raisebox{0.15em}{\includegraphics[width=0.02\linewidth]{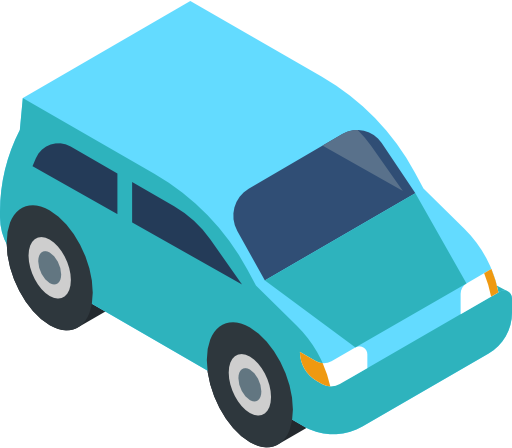}}}
\author[]{Steven~C.~H.~Hoi$^{4,5}$}

\affiliation[]{$^1$Zhejiang University \quad
$^2$National University of Singapore\quad
$^3$Nanjing University of Aeronautics and Astronautics\quad
$^4$Alibaba Group\quad 
$^5$Singapore Management University\\[1.8ex]
\raisebox{-0.1em}{\includegraphics[width=0.029\linewidth]{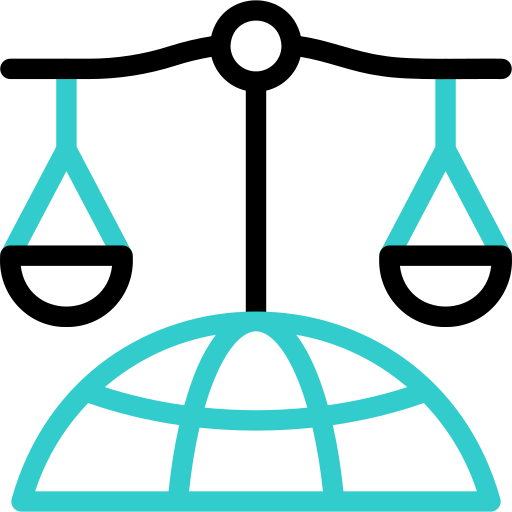}}~WorldBench Team
\\[2.2ex]
\raisebox{-0.2em}{\includegraphics[width=0.032\linewidth]{figs/icons/car1.png}}~{\small \textbf{Equal Contributions}}
\quad
\raisebox{-0.2em}{\includegraphics[width=0.028\linewidth]{figs/icons/car2.png}}~{\small \textbf{Corresponding Author}}
}

\abstract{
The rapid advancement of autonomous systems, including self-driving vehicles and drones, has intensified the need to forge true Spatial Intelligence from multi-modal onboard sensor data. While foundation models excel in single-modal contexts, integrating their capabilities across diverse sensors like cameras and LiDAR to create a unified understanding remains a formidable challenge. This paper presents a comprehensive framework for multi-modal pre-training, identifying the core set of techniques driving progress toward this goal. We dissect the interplay between foundational sensor characteristics and learning strategies, evaluating the role of platform-specific datasets in enabling these advancements. Our central contribution is the formulation of a unified taxonomy for pre-training paradigms: ranging from single-modality baselines to sophisticated unified frameworks that learn holistic representations for advanced tasks like 3D object detection and semantic occupancy prediction. Furthermore, we investigate the integration of textual inputs and occupancy representations to facilitate open-world perception and planning. Finally, we identify critical bottlenecks, such as computational efficiency and model scalability, and propose a roadmap toward general-purpose multi-modal foundation models capable of achieving robust Spatial Intelligence for real-world deployment.
}

\metadata[
\raisebox{-0.2em}{\includegraphics[width=0.025\linewidth]{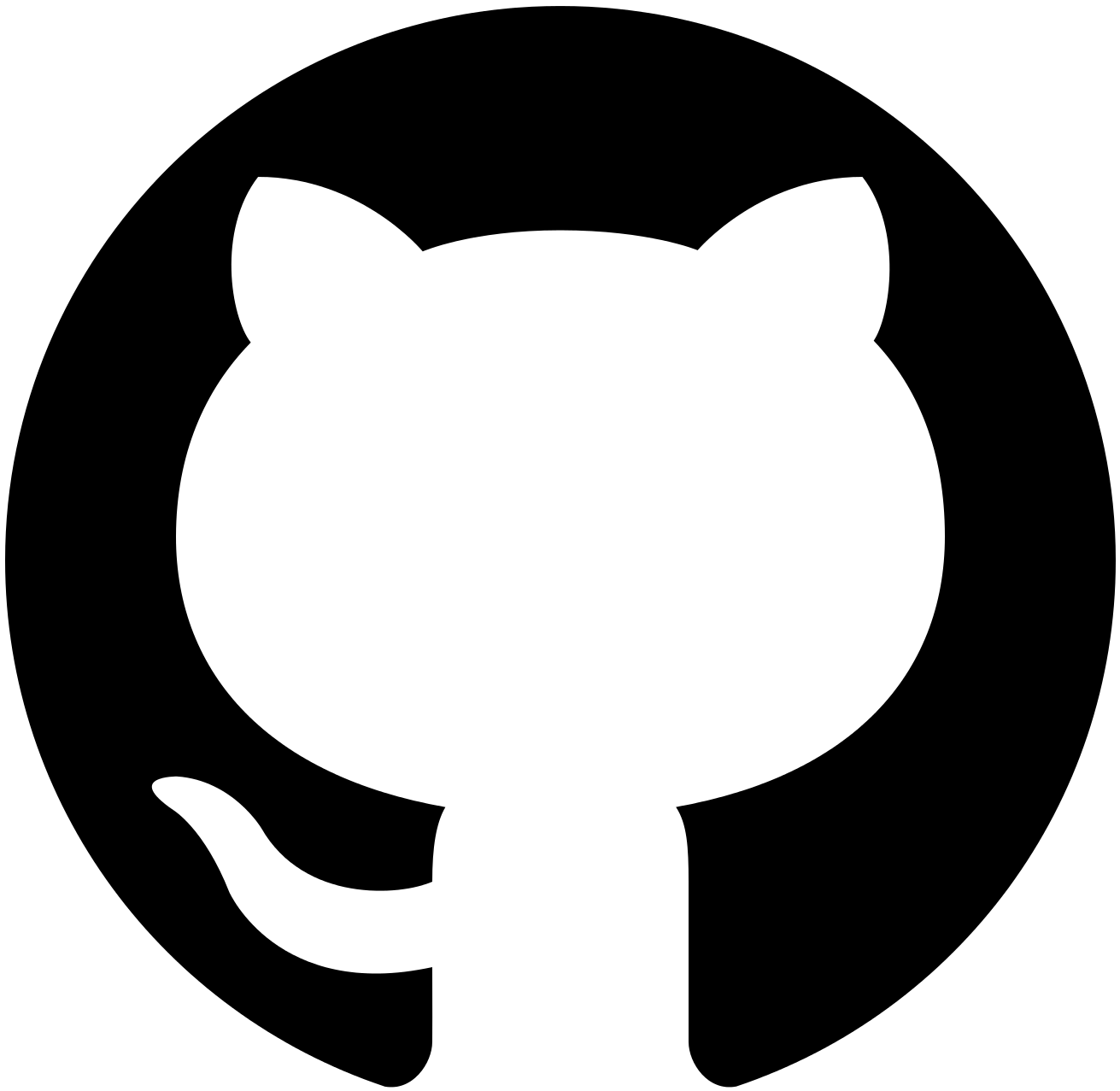}}~~GitHub Repo]{\href{https://github.com/worldbench/awesome-spatial-intelligence}{\texttt{https://github.com/worldbench/awesome-spatial-intelligence}}
\\[-1ex]}

\begin{document}

\maketitle

\section{Introduction}
\label{sec:introduction}

With the rapid proliferation of autonomous platforms, ranging from self-driving vehicles~\cite{hu2023planning, li2022bevformer, min2024driveworld} and aerial drones~\cite{meier2024cdrone, zhao2024biodrone} to unmanned surface vehicles~\cite{yao2024waterscenes, guan2024watervg}, rail-based systems~\cite{zendel2019railsem19, li2022rail, zouaoui2022railset}, and legged robots~\cite{han2024lifelike, ha2024learning}, the challenge of endowing machines with the capability to perceive and act in the real world has reached an unprecedented level of complexity. 
As these systems are required to navigate diverse and dynamically evolving scenarios, they demand a robust and deeply nuanced understanding of their environment to support critical downstream functions, including navigation~\cite{maaref2002sensor, alqobali2023survey}, interaction~\cite{jevtic2015comparison}, and planning~\cite{hu2023planning, liu2025occvla}.

Central to these platforms is a sophisticated suite of onboard sensors, primarily comprising cameras, LiDAR, radar, and emerging event cameras, which collectively serve as the foundation for perception~\cite{geiger2012we, caesar2020nuscenes, hu2021towards, chaney2023m3ed}. 
Each modality contributes a unique and complementary stream of information, where cameras provide rich visual semantics~\cite{xie2023robobev, li2022bevformer}, LiDAR delivers precise 3D geometry~\cite{liang2025lidarcrafter, krispel2024maeli}, and radar captures essential motion cues~\cite{pushkareva2024radar, yao2023radar}, while event cameras offer microsecond-level temporal precision for high-speed dynamics~\cite{yang2023event, kong2025eventfly, han2025e-deflare, lu2025flexevent}. 
The effective integration of these heterogeneous data streams is paramount for achieving the holistic perception required for safe and generalizable autonomy~\cite{sautier2022image, xie2025benchmarking, kong2023robo3d, place3d, he2019rethinking}.
In response to this demand, the research community has curated numerous large-scale and sensor-centric datasets~\cite{geiger2012we,sun2020scalability, Argoverse}, alongside specialized benchmarks for drones~\cite{zhao2024biodrone, meier2024cdrone} and robotic agents~\cite{han2024lifelike}. 

While these datasets provide an invaluable empirical foundation, they simultaneously highlight a fundamental challenge that constitutes the central focus of this work. 
Most existing datasets heavily rely on costly manual annotations to support supervised learning paradigms~\cite{geiger2012we, caesar2020nuscenes, guan2024watervg}, creating significant bottlenecks for scalability and generalization~\cite{tian2024occ3d, wang2024label, kong2023rethinking, xie2025drivebench, kong2023laserMix}.
Consequently, there has been a growing interest in representation learning methods that aim to distill meaningful features directly from raw sensor data, alleviating the dependence on extensive human supervision~\cite{he2020momentum, he2022masked, xiao2023unsupervised, hegde2024equivariant}. 
Particularly noteworthy is the emergence of foundation models, which facilitate large-scale, transferable pre-training across various domains, including vision~\cite{radford2021clip, kirillov2023sam}, 3D geometry~\cite{liu2023geomim, zhang2024bevworld}, and multi-modal scenarios~\cite{yang2024unipad, zou2024unim, kong2025multi}. 
Such foundation models provide a unified paradigm for extracting general-purpose representations from diverse sensor inputs~\cite{mahmoud2024image, yang2023vidar, kong2023robodepth}, significantly enhancing cross-domain adaptability and paving the way for next-generation world models~\cite{kong20253d, min2024driveworld, zheng2023occworld, survey_vla4ad, worldlens, liu2025lalalidar, liang2025lidarcrafter, xu2025U4D}. 

\begin{figure}[t]
\begin{center}
\includegraphics[width=\linewidth]{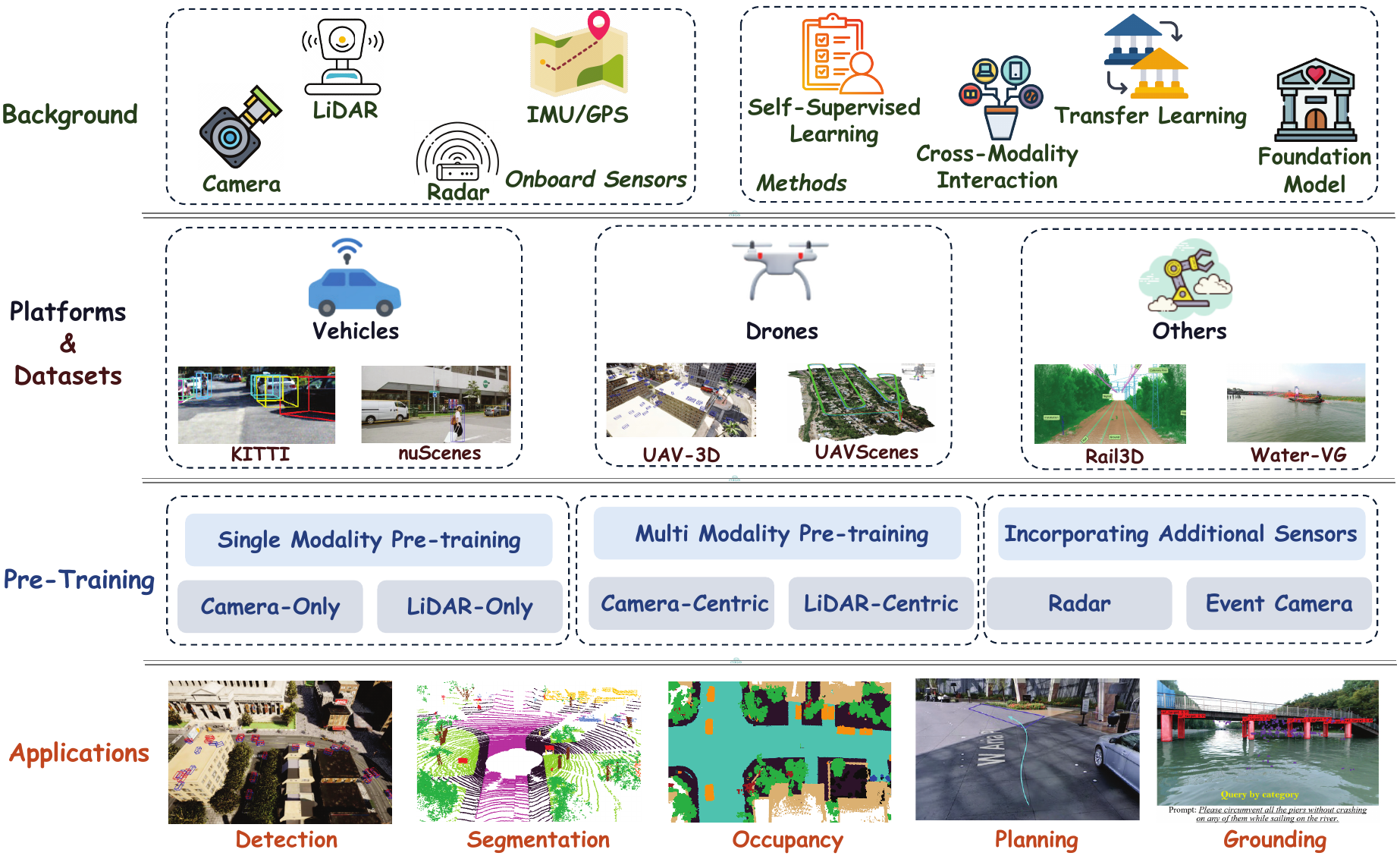} 
\end{center}
\vspace{-1mm}
\caption{\textbf{Overview of the paper structure.} We systematically structure the landscape of multi-modal data pre-training for forging Spatial Intelligence. This work is organized into four key pillars: (1) \textbf{Background}, introducing onboard sensors and foundational learning paradigms; (2) \textbf{Platforms \& Datasets}, analyzing benchmarks across autonomous vehicles, drones, and other robotic systems; (3) \textbf{Pre-Training Methodologies}, categorized into single-modality, cross-modal (Camera/LiDAR-centric), and unified frameworks; and (4) \textbf{Applications}, highlighting downstream tasks from 3D perception to open-world planning.}
\label{fig:overallnetwork}
\end{figure}

As a result, pre-training strategies tailored specifically to sensor modalities have become an essential research frontier. As depicted in Fig.~\ref{fig:overallnetwork}, these strategies form the core techniques to forge what we define as \textbf{Spatial Intelligence} -- a capability that transcends simple detection to encompass holistic scene understanding, reasoning, and future prediction~\cite{liu2025occvla, yang2023vidar, liang2025seeing}. Current approaches include single-modality methods (\emph{e.g.}, LiDAR-only or camera-only)~\cite{krispel2024maeli, agro2024uno, hindel2023inod, zhang2024visionpad}, cross-modal knowledge transfer (\emph{e.g.}, distillation between camera and LiDAR)~\cite{yang2023vidar,sautier2022image, kong2025largead}, and unified multi-modal pre-training frameworks~\cite{yang2024unipad, zou2024unim}. Understanding the overall landscape of these methods, as well as their connections to sensor characteristics, platform constraints, and the development of foundation models, is crucial for advancing robust and efficient perception capabilities in intelligent systems.

In this work, we systematically analyze the state-of-the-art techniques in representation learning from onboard sensor data, emphasizing multi-modal interactions and integration with foundation models. 
We first dissect foundational methodologies such as self-supervised learning, transfer learning, and multimodal learning, evaluating their respective strengths and limitations across various autonomous platforms, including self-driving vehicles, drones, robotic dogs, and rail transportation systems. By structuring and characterizing representative pre-training approaches according to modality composition, sensor interactions, and targeted applications, we highlight their adaptability to diverse sensor configurations and practical scenarios. Furthermore, we investigate key challenges in sensor representation learning, such as data sparsity, sensor noise, multi-modal alignment, and real-time processing demands. Finally, we propose promising directions for future research toward generative world models and embodied reasoning suitable for dynamic real-world environments.

\subsection{Scope of the Work}

Multi-modal representation learning from onboard sensors encompasses various related areas, such as single-modality pre-training, cross-modal fusion, and foundation model integration. Given the breadth of these topics, it is impractical to exhaustively analyze all relevant methods within a single manuscript. 
Therefore, this work specifically concentrates on recent advances in foundation models for multi-modal representation learning, primarily focusing on onboard camera and LiDAR sensors~\cite{yang2024unipad, yang2023vidar, shi2025drivex}. Representative methods involving additional sensors such as radar and event cameras are also examined~\cite{yao2023radar, dong2021radar, yang2023event, yang2024event}.

We emphasize significant progress from the past five years, particularly highlighting influential works published in top-tier conferences and journals. In addition to technical approaches, we analyze widely adopted datasets, evaluation metrics, and sensor configurations. Finally, we investigate key challenges and outline promising future research directions.

\subsection{Relation to Previous Studies}

Several existing studies~\cite{xiao2023unsupervised, yan2024forging, zong2023self, zhu2024sora, wu2025foundation} have recently explored representation learning in autonomous systems, typically focusing on individual sensor modalities, specific autonomous platforms, or particular downstream tasks. While these efforts offer valuable insights into targeted aspects of the field, they often discuss sensors and tasks independently, lacking an integrated perspective on how single-modality and cross-modal approaches collectively advance multi-modal representation learning.

In contrast, our work presents a comprehensive framework \textbf{emphasizing the role of foundation models within multi-modal representation learning from onboard sensors}. We systematically analyze modality-specific pre-training as well as cross-modal interactions and unified frameworks, clearly highlighting how these methods interconnect and contribute to robust, generalizable perception across various platforms and tasks. By bridging single-modality strategies with unified multi-modal paradigms, this study uniquely facilitates an in-depth understanding of recent advances, emerging trends, and future directions in multi-modal pre-training for autonomous systems.

\subsection{Organization}
To provide a clear roadmap of the field, we present a comprehensive taxonomy encompassing datasets, pre-training paradigms, and downstream applications, as illustrated in Fig.~\ref{fig:taxonomy}.
The remainder of this paper is structured as follows:
\begin{itemize}
    \item Section~\ref{sec:bg} introduces the foundations of data representation learning for onboard sensors, covering sensor characteristics, pre-training paradigms, and the role of foundation models.

    \item Section~\ref{sec:datasets} analyzes platform-specific datasets, including those collected from autonomous vehicles, aerial drones, and other robotic systems.

    \item Section~\ref{sec:methods} provides a comprehensive analysis of pre-training methods, categorized by sensor modality, interaction level, and downstream application.

    \item Section~\ref{sec:open} investigates recent progress in open-world perception and planning, focusing on text-assisted understanding and the shift toward generative world models for end-to-end autonomy.

    \item Section~\ref{sec:chall_and_future} outlines key challenges in current research and highlights promising future directions.

    \item Section~\ref{sec:conc} concludes the paper with a summary of major insights and takeaways.
\end{itemize}
\section{Background}
\label{sec:bg}

Multi-modal pre-training from onboard sensors serves as the bedrock for forging \textbf{Spatial Intelligence} in autonomous systems. 
It aims to transcend simple feature extraction by distilling compact, discriminative, and semantically rich representations from diverse sensory inputs.
By effectively integrating complementary information from modalities such as cameras, LiDAR and radar, these methods enable foundation models to not only perceive geometry and semantics but also reason about dynamics and affordances~\cite{yang2023vidar, liang2025seeing}.
In the context of autonomous deployment, developing scalable and reliable multi-modal pre-training approaches is essential for achieving robust open-world generalization and bridging the gap between passive perception and active embodied reasoning.

\subsection{Onboard Sensors and Data Characteristics}

The sensory apparatus of intelligent agents, primarily comprising cameras, LiDAR, radar, and event cameras, presents a heterogeneous data landscape characterized by distinct modalities and formats. 
Cameras provide dense semantic and textural information essential for scene understanding~\cite{yang2023vidar, li2022bevformer}, yet they remain susceptible to environmental variations such as illumination changes and adverse weather conditions. 
In contrast, LiDAR sensors deliver precise 3D geometric structures via point clouds~\cite{xie2020pointcontrast, krispel2024maeli, mahmoud2023self}, which offer robustness against lighting variations but suffer from inherent sparsity and limited semantic richness. 
Radar provides robust Doppler velocity cues even in adverse weather, albeit at lower spatial resolution~\cite{yao2023radar,dong2021radar}. Complementing these with microsecond-level temporal precision, event cameras capture asynchronous brightness changes to handle high-speed dynamics and motion blur inherent in standard vision~\cite{yang2023event, kong2025eventfly,zhou2024eventbind,kong2025talk2event,kong2024openess}.
Understanding the inherent properties of these sensors, specifically the trade-off between the \emph{semantic richness} of vision and the \emph{geometric precision} of ranging sensors, is fundamental. 
Effective representation learning must address these disparities to construct a unified and coherent world model.

\subsection{Paradigms of Representation Learning}
\label{sec:data-representation-learning}

\begin{figure}[t]
\begin{center}
\includegraphics[width=\linewidth]{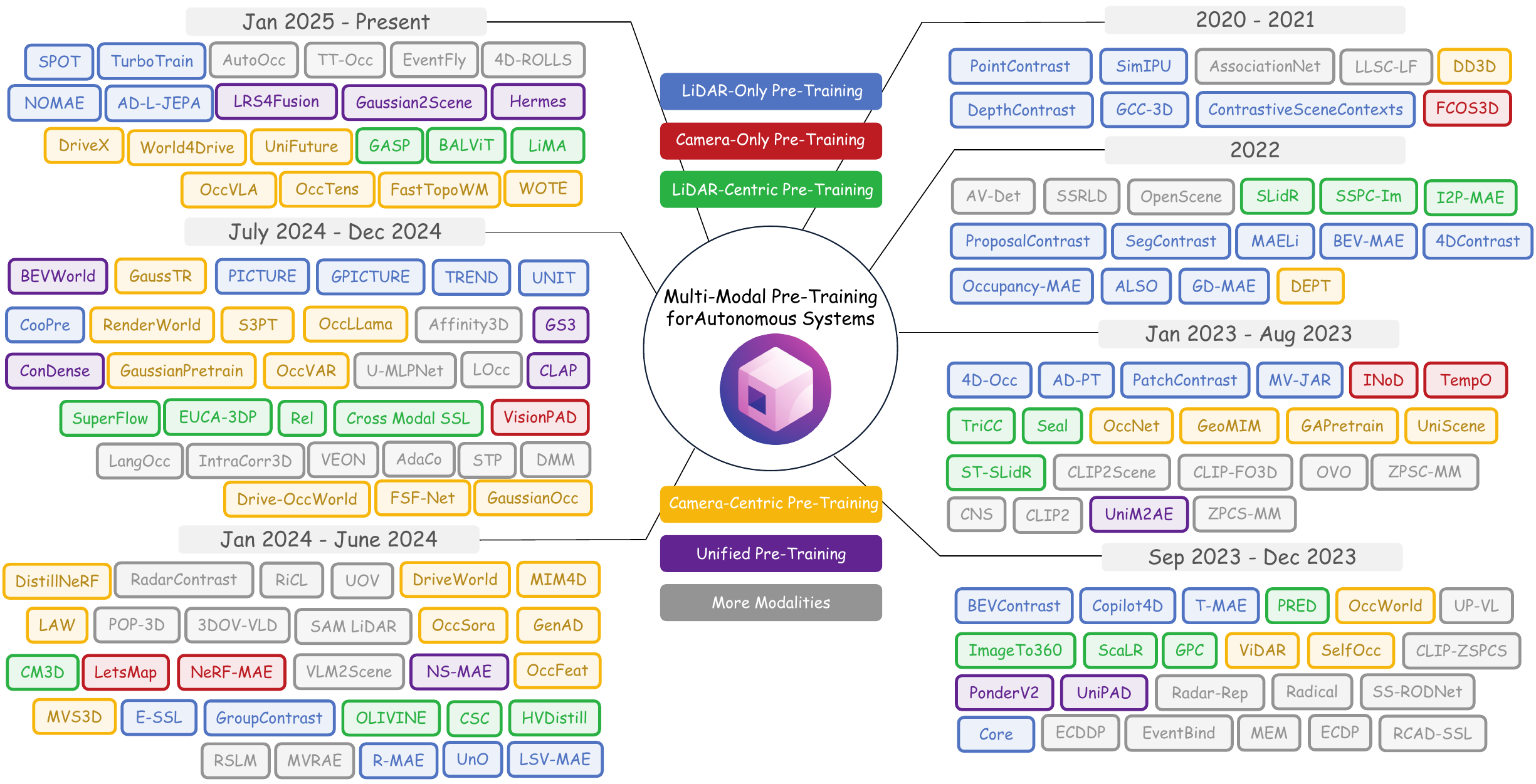} 
\end{center}
\vspace{-1mm}
\caption{\textbf{Chronological evolution of representative pre-training methods (2020--2025).} The timeline illustrates the paradigm shift in representation learning for autonomous systems. Early approaches predominantly focused on single-modality self-supervision (\eg, \emph{LiDAR-only} contrastive learning). In contrast, recent advancements (2023--present) demonstrate a surge in cross-modal synergy, characterized by \emph{Camera/LiDAR-centric} methods and \emph{Unified} pre-training frameworks, ultimately paving the way for generative world models and comprehensive spatial intelligence.}
\label{fig:timeline}
\end{figure}

To forge spatial intelligence from the heterogeneous data streams described above, a robust methodological framework is required. 
The evolution of this framework has followed a clear logical progression, as chronologically illustrated in Fig.~\ref{fig:timeline}. Initially, to overcome the immense cost of manual annotation, the field turned to \textbf{Self-Supervised Learning} to extract meaningful features directly from vast quantities of unlabeled data. 
A natural next step was to exploit the complementary nature of different sensors through \textbf{Cross-Modality Interaction}, creating a more holistic representation than any single sensor could provide. 
Concurrently, \textbf{Knowledge Distillation and Transfer Learning} emerged as a powerful technique to leverage priors from well-established vision foundation models, accelerating progress in 3D domains. 
Ultimately, these distinct paradigms are being synthesized under a unified vision: the development of \textbf{Foundation Models} and \textbf{Generative World Models}.

\subsubsection{Self-Supervised Learning}
Self-supervised learning (SSL) has emerged as the dominant paradigm for representation learning from unlabeled sensor data~\cite{he2020momentum, he2022masked, caron2021dino, oquabdinov2}. 
By defining suitable pretext tasks, models leverage supervisory signals inherently present within the data.
Classic strategies include \emph{Contrastive Learning}, which discriminates between augmented views of the same instance, and \emph{Masked Modeling}, which reconstructs obscured portions of inputs~\cite{he2022masked, krispel2024maeli, klenk2024masked}.
More recently, \textbf{Generative Modeling} (\eg, next-token prediction or video generation) has gained prominence~\cite{bartoccioni2025vavim, zheng2023occworld}. 
By learning to predict future frames or occupancy states, these methods enable models to internalize the physics and dynamics of the environment, serving as a precursor to world models.

\subsubsection{Cross-Modality Interaction}
Cross-modal interaction methods aim to fuse disparate sensor modalities into a unified representation space, enhancing both robustness and semantic depth~\cite{yang2023vidar, min2024driveworld, shi2025drivex, xu2025beyond}.
For instance, projecting dense visual features from cameras onto sparse LiDAR point clouds allows models to achieve superior spatial-semantic reasoning~\cite{yin2021multimodal, xu2021fusionpainting, bai2022transfusion}.
Key challenges addressed by these approaches include spatio-temporal alignment, handling modality-specific noise, and maintaining robustness when one modality is degraded or missing.

\subsubsection{Knowledge Distillation and Transfer Learning}
While transfer learning traditionally involves adapting pre-trained weights to new domains~\cite{yosinski2014transferable, liu2024segment}, in the context of multi-modal autonomous systems, it increasingly takes the form of \textbf{Knowledge Distillation}.
Here, powerful 2D vision foundation models (\emph{teachers}) are used to guide the training of 3D sensor backbones (\emph{students}). 
This allows 3D models to inherit the open-vocabulary capabilities and rich semantics of large-scale vision models~\cite{radford2021clip, kirillov2023sam, oquabdinov2} without requiring massive annotated 3D datasets~\cite{sautier2022image, mahmoud2024image}.
This paradigm effectively bridges the data scale gap between the 2D image domain and the 3D robotics domain.

\subsubsection{Foundation Models}
\label{sec:foundation-models}

Foundation models represent a paradigm shift from specialized pipelines to unified, scalable representation learning~\cite{bolya2025perception, simeoni2025dinov3}.
In the vision domain, the trajectory from CNNs~\cite{he2016resnet} to Transformers~\cite{dosovitskiy2021vit} and general-purpose encoders like DINO~\cite{oquabdinov2, caron2021dino} and SAM~\cite{kirillov2023sam, ravi2024sam} has established robust, transferable perceptual priors.
Recent research integrates these visual priors into non-visual modalities (\eg, LiDAR and radar) via cross-modal alignment, enriching 3D perception with open-world semantics~\cite{kong2025largead, yao2023radar, xu2025beyond}.
Crucially, the field is now advancing beyond perception towards \textbf{Generative World Models}~\cite{zheng2023occworld, kong20253d} and \textbf{Vision-Language-Action (VLA)} models~\cite{driess2023palm, wang2024qwen2, liu2025occvla}. 
These next-generation foundation models integrate vision, language, and action into a unified reasoning framework~\cite{survey_vla4ad, li2025drivevla}, enabling systems not just to recognize objects, but to simulate future scenarios and plan actions in complex, dynamic environments.
\tikzstyle{my-box}=[
    rectangle,
    draw=hidden-draw,
    rounded corners,
    text opacity=1,
    minimum height=1.5em,
    minimum width=5em,
    inner sep=2pt,
    align=center,
    fill opacity=.5,
    line width=0.8pt,
]
\tikzstyle{leaf}=[my-box, minimum height=1.5em,
    fill=hidden-pink!80, text=black, align=left,font=\normalsize,
    inner xsep=2pt,
    inner ysep=4pt,
    line width=0.8pt,
]

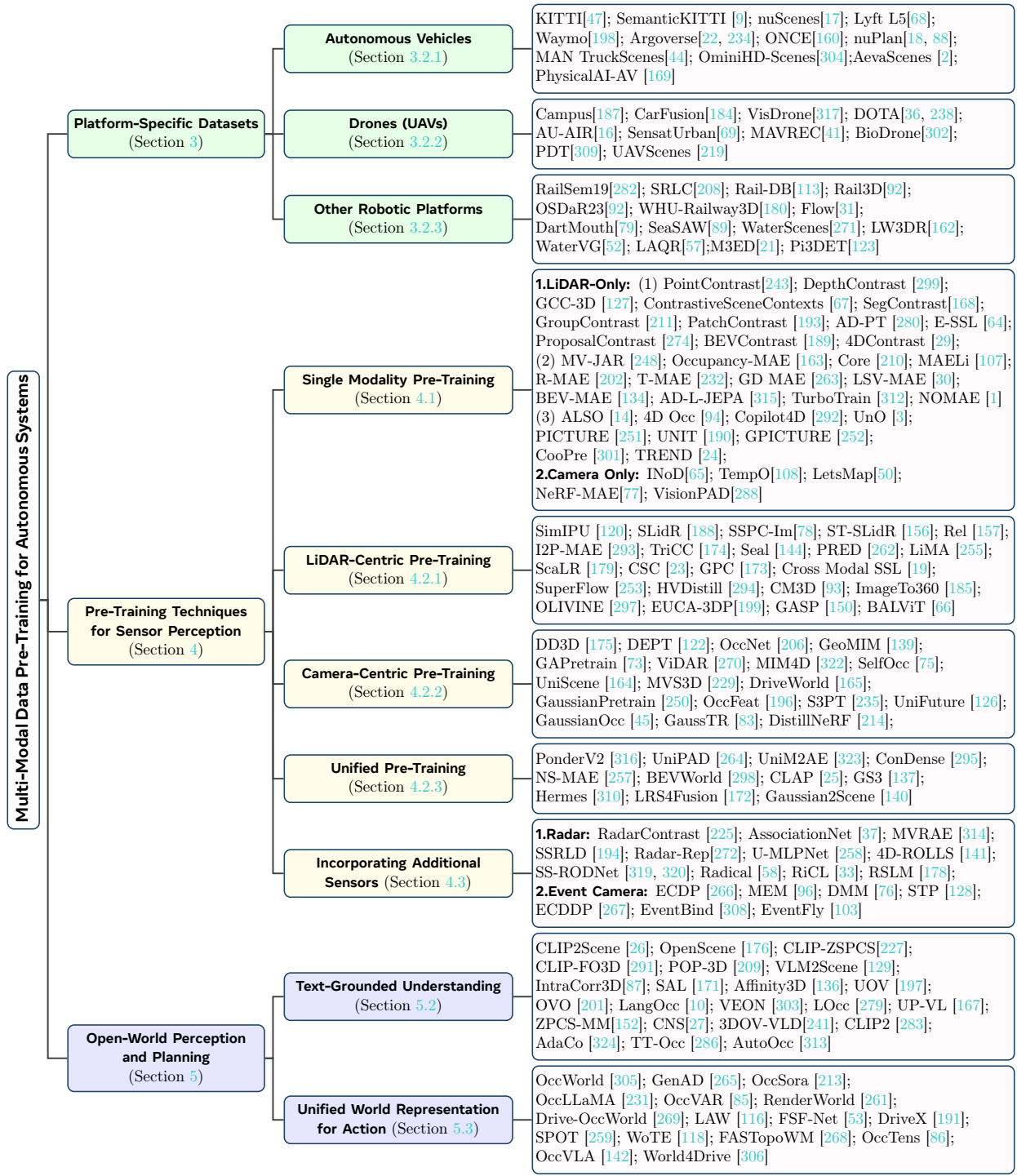
\begin{figure*}[]
    \centering
    \resizebox{\textwidth}{!}{
        \begin{forest}
            forked edges,
            for tree={
                grow=east,
                reversed=true,
                anchor=base west,
                parent anchor=east,
                child anchor=west,
                base=center,
                font=\large,
                rectangle,
                draw=hidden-draw,
                rounded corners,
                align=left,
                text centered,
                minimum width=4em,
                edge+={darkgray, line width=1pt},
                s sep=3pt,
                inner xsep=2pt,
                inner ysep=3pt,
                line width=0.8pt,
                ver/.style={rotate=90, child anchor=north, parent anchor=south, anchor=center},
            },
            where level=1{text width=12em,align=center,font=\normalsize,}{},
            where level=2{text width=14em,align=center,font=\normalsize,}{},
            where level=3{text width=16em,font=\normalsize,}{},
            where level=4{text width=35em,font=\normalsize,}{},
            where level=5{text width=18em,font=\normalsize,}{},
            [
                \textbf{~Multi-Modal Data Pre-Training for Autonomous Systems~}, ver
                [
                        \textbf{Platform-Specific Datasets} \\  (Section~\ref{sec:datasets}),
                        fill=green!10
                        [
                            \textbf{Autonomous Vehicles} \\ (Section~\ref{sec:dataset_vehicle}), fill=green!10
                            [
                                KITTI\cite{geiger2012we};
                                SemanticKITTI~\cite{behley2019iccv};
                                nuScenes\cite{caesar2020nuscenes}; 
                                Lyft L5\cite{houston2020one};\\
                                Waymo\cite{sun2020scalability};
                                Argoverse\cite{Argoverse, Argoverse2};
                                ONCE\cite{mao2021one};
                                nuPlan\cite{caesar2021nuplan, karnchanachari2024towards};\\
                                MAN TruckScenes\cite{fent2024man};
                                OminiHD-Scenes\cite{zheng2024omnihd};AevaScenes~\cite{aevascenes2025}; \\
                                PhysicalAI-AV~\cite{nvidia_physicalai_av_2025}
                                ,leaf, text width=30em
                            ]
                        ]
                        [
                            \textbf{Drones (UAVs)} \\ (Section~\ref{sec:dataset_drone}), fill=green!10
                            [
                                Campus\cite{robicquet2016learning};
                                CarFusion\cite{reddy2018carfusion};
                                VisDrone\cite{zhu2021detection};
                                DOTA\cite{xia2018dota,ding2021object};\\
                                AU-AIR\cite{bozcan2020air};
                                SensatUrban\cite{hu2021towards};
                                MAVREC\cite{dutta2024multiview};
                                BioDrone\cite{zhao2024biodrone};\\
                                PDT\cite{zhou2025pdt}; UAVScenes~\cite{wang2025uavscenes}
                                ,leaf, text width=30em
                            ]
                        ]
                        [
                            \textbf{Other Robotic Platforms} \\ (Section~\ref{sec:dataset_others}), fill=green!10
                              [
                                RailSem19\cite{zendel2019railsem19};
                                SRLC\cite{uggla2021towards};
                                Rail-DB\cite{li2022rail};
                                Rail3D\cite{kharroubi2024multi};\\
                                OSDaR23\cite{kharroubi2024multi};
                                WHU-Railway3D\cite{qiu2024whu};
                                Flow\cite{cheng2021flow};\\
                                DartMouth\cite{jeong2021efficient};
                                SeaSAW\cite{kaur2022sea};
                                WaterScenes\cite{yao2024waterscenes};
                                LW3DR\cite{miki2024learning};\\
                                WaterVG\cite{guan2024watervg};
                                LAQR\cite{han2024lifelike};M3ED\cite{chaney2023m3ed}; Pi3DET\cite{liang2025perspective}
                                ,leaf, text width=30em
                            ]
                        ]
                    ]
                     [
                        \textbf{Pre-Training Techniques} \\ \textbf{for Sensor Perception}  \\ (Section~\ref{sec:methods}), fill=yellow!10
                        [
                            \textbf{Single Modality Pre-Training} \\ (Section~\ref{sec:single}), fill=yellow!10
                            [
                            \textbf{1.LiDAR-Only:} (1) PointContrast\cite{xie2020pointcontrast}; DepthContrast~\cite{zhang2021self}; \\ 
                            GCC-3D~\cite{liang2021exploring}; ContrastiveSceneContexts~\cite{hou2021exploring}; SegContrast\cite{nunes2022segcontrast}; \\GroupContrast~\cite{wang2024groupcontrast}; PatchContrast~\cite{shrout2023patchcontrast}; AD-PT~\cite{yuan2024ad}; E-SSL~\cite{hegde2024equivariant};\\ ProposalContrast~\cite{yin2022proposalcontrast}; BEVContrast~\cite{sautier2024bevcontrast}; 4DContrast~\cite{chen20224dcontrast};\\ (2) MV-JAR~\cite{xu2023mv}; Occupancy-MAE~\cite{min2023occupancy}; Core~\cite{wang2023core}; MAELi~\cite{krispel2024maeli}; \\R-MAE~\cite{tayebati2024sense};  T-MAE~\cite{wei2024t}; GD MAE~\cite{yang2023gd}; LSV-MAE~\cite{cheng2024rethinking}; \\  BEV-MAE~\cite{lin2024bev}; AD-L-JEPA~\cite{zhu2025adljepa}; TurboTrain~\cite{zhou2025turbotrain}; NOMAE~\cite{abdelsamad2025nomae} \\ (3) ALSO~\cite{boulch2023also}; 4D Occ~\cite{khurana2023point}; Copilot4D~\cite{zhangcopilot4d};   UnO~\cite{agro2024uno};  \\ PICTURE~\cite{xu2024point};   UNIT~\cite{sautier2024unit};   GPICTURE~\cite{xu2024mutual}; \\ CooPre~\cite{zhao2024coopre};  TREND~\cite{chen2024trend};  \\
                            \textbf{2.Camera Only:} 
                            INoD\cite{hindel2023inod};
                            TempO\cite{lang2024self};
                            LetsMap\cite{gosala2025letsmap}; \\
                            NeRF-MAE\cite{irshad2025nerf};
                            VisionPAD\cite{zhang2024visionpad}\\
                             , leaf, text width=30em
                            ]
                        ]
                        [
                            \textbf{LiDAR-Centric Pre-Training} \\ (Section~\ref{sec:lidar_centric}), fill=yellow!10
                            [
                                SimIPU~\cite{li2022simipu}; SLidR~\cite{sautier2022image}; SSPC-Im\cite{janda2022self}; ST-SLidR~\cite{mahmoud2023self}; Rel~\cite{mahmoud2024image}; \\ 
                                I2P-MAE~\cite{zhang2023learning}; TriCC~\cite{pang2023unsupervised}; Seal~\cite{liu2024segment}; PRED~\cite{yang2024pred}; 
                                LiMA~\cite{xu2025beyond}; \\ ScaLR~\cite{puy2024three};  CSC~\cite{chen2024building}; GPC~\cite{panpre}; Cross Modal SSL~\cite{cai2024cross}; \\ SuperFlow~\cite{xu20244d};  HVDistill~\cite{zhang2024hvdistill}; CM3D~\cite{khuranashelf}; ImageTo360~\cite{reichardt2023360};\\ OLIVINE~\cite{zhang2024fine}; EUCA-3DP\cite{sun2024exploring}; GASP~\cite{ljungbergh2025gasp}; BALViT~\cite{hindel2025BALViT}
                                ,leaf, text width=30em
                            ]
                        ]
                        [
                            \textbf{Camera-Centric Pre-Training} \\ (Section~\ref{sec:camera_centric}), fill=yellow!10
                            [
                            DD3D~\cite{park2021pseudo}; DEPT~\cite{li2022delving}; OccNet~\cite{tong2023scene}; GeoMIM~\cite{liu2023geomim}; \\ GAPretrain~\cite{huang2025lidar}; ViDAR~\cite{yang2023vidar}; MIM4D~\cite{zou2024mim4d}; SelfOcc~\cite{huang2024selfocc}; \\  UniScene~\cite{min2024multi}; MVS3D~\cite{wang2024focus}; DriveWorld~\cite{min2024driveworld}; \\ GaussianPretrain~\cite{xu2024gaussianpretrain}; 
                                OccFeat~\cite{sirko2024occfeat};
                                S3PT~\cite{wozniak2024s3pt}; UniFuture~\cite{liang2025seeing};  \\ GaussianOcc~\cite{gan2024gaussianocc}; GaussTR~\cite{jiang2024gausstr}; DistillNeRF~\cite{wang2024distillnerf}; 
                                ,leaf, text width=30em
                            ]
                        ]
                        [
                            \textbf{Unified Pre-Training} \\ (Section~\ref{sec:unified}), fill=yellow!10
                            [
                                PonderV2~\cite{zhu2023ponderv2}; UniPAD~\cite{yang2024unipad}; UniM2AE~\cite{zou2024unim}; ConDense~\cite{zhang2024condense}; \\
                                NS-MAE~\cite{xu2024learningshared}; BEVWorld~\cite{zhang2024bevworld}; CLAP~\cite{chen2024clap}; GS3~\cite{liu2024point}; \\Hermes~\cite{zhou2025hermes}; LRS4Fusion~\cite{palladin2025self}; Gaussian2Scene~\cite{liu2025gaussian2scene},leaf, text width=30em
                            ]
                        ]
                        [
                            \textbf{Incorporating Additional} \\ \textbf{Sensors} (Section~\ref{sec:other_sensors}), fill=yellow!10
                            [
                                \textbf{1.Radar:} RadarContrast~\cite{wang2024self}; AssociationNet~\cite{dong2021radar}; MVRAE~\cite{zhu2024multi}; \\ SSRLD~\cite{si2022self};  Radar-Rep\cite{yao2023radar};  U-MLPNet~\cite{yan2024learning};  4D-ROLLS~\cite{liu20254d}; \\ SS-RODNet~\cite{zhuang2023effective, zhuang2023pre}; Radical~\cite{hao2024bootstrapping}; RiCL~\cite{decourt2024leveraging}; RSLM~\cite{pushkareva2024radar};\\
                                \textbf{2.Event Camera:} ECDP~\cite{yang2023event}; MEM~\cite{klenk2024masked}; DMM~\cite{huang2024data};  STP~\cite{liangenhancing}; \\  ECDDP~\cite{yang2024event}; EventBind~\cite{zhou2024eventbind}; EventFly~\cite{kong2025eventfly}
                                ,leaf, text width=30em
                            ]   
                        ]
                    ]
                    [
                        \textbf{Open-World Perception} \\ \textbf{and Planning} \\ (Section~\ref{sec:open}), fill=blue!10
                        [
                            \textbf{Text-Grounded Understanding} \\  (Section~\ref{sec:text_ground}), fill=blue!10
                            [
                                CLIP2Scene~\cite{chen2023clip2scene}; OpenScene~\cite{peng2023openscene}; CLIP-ZSPCS\cite{wang2023transferring}; \\ CLIP-FO3D~\cite{zhang2023clip};
                                POP-3D~\cite{vobecky2024pop}; VLM2Scene~\cite{liao2024vlm2scene}; \\ IntraCorr3D\cite{kang2024hierarchical}; SAL~\cite{ovsep2024better}; 
                                Affinity3D~\cite{liu2024affinity3d}; UOV~\cite{sun20243d}; \\ OVO~\cite{tan2023ovo}; LangOcc~\cite{boeder2024langocc}; VEON~\cite{zheng2024veon};  LOcc~\cite{yu2024language}; UP-VL~\cite{najibi2023unsupervised}; \\ ZPCS-MM\cite{lu2023see}; CNS\cite{chen2024towards}; 3DOV-VLD\cite{xiao20253d}; CLIP2~\cite{zeng2023clip2}; \\ AdaCo~\cite{zou2024adaco}; TT-Occ~\cite{zhang2025tt}; AutoOcc~\cite{zhou2025autoocc} 
                                ,leaf, text width=30em
                            ] 
                        ]
                        [
                            \textbf{Unified World Representation} \\ \textbf{for Action} (Section~\ref{sec:world_model}), fill=blue!10
                            [
                            OccWorld~\cite{zheng2023occworld}; GenAD~\cite{yang2024generalized}; OccSora~\cite{wang2024occsora}; \\ OccLLaMA~\cite{wei2024occllama};  OccVAR~\cite{jinoccvar};   RenderWorld~\cite{yan2024renderworld}; \\ Drive-OccWorld~\cite{yang2024driving}; LAW~\cite{li2024enhancing}; FSF-Net~\cite{guo2024fsf}; DriveX~\cite{shi2025drivex}; \\ SPOT~\cite{yan2025spot}; WoTE~\cite{li2025end}; FASTopoWM~\cite{yang2025fastopowm}; OccTens~\cite{jin2025occtens}; \\OccVLA~\cite{liu2025occvla}; World4Drive~\cite{zheng2025world4drive}
                            , leaf, text width=30em
                            ]
                        ]
                    ]
            ]
        \end{forest}
      }
    \caption{\textbf{Taxonomy of multi-modal pre-training methodologies.} We structure the landscape into three pillars: (1) Platform-specific datasets, (2) Core pre-training techniques classified by sensor interaction (single-modality, cross-modal, and unified), and (3) Advanced open-world perception and planning tasks.}
    \label{fig:taxonomy}
\end{figure*}
\begin{table}[t]
    \centering
    \renewcommand\arraystretch{1.2}
    \setlength\tabcolsep{0.1cm}
    \caption{\textbf{Summary of representative autonomous vehicle datasets.} 
\textbf{Region}: ``AS'' (Asia), ``EU'' (Europe), ``NA'' (North America). 
\textbf{Sensor Configuration}: ``Camera'', ``LiDAR'', and ``Radar'' denote the count of equipped sensors. 
\textbf{Data Statistics}: \textbf{Scenes} refers to the number of dataset clips/sequences; \textbf{Frames} indicates the total annotated frames. 
\textbf{Conditions}: \textbf{Weather} captures adverse scenarios; ``d\&n'' denotes day and night coverage. 
Symbol ``-'' indicates that the specific modality or statistic is unavailable/unsupported.}
    \vspace{-0.2cm}
    \resizebox{\textwidth}{!}{
    \begin{tabular}{r|c|c|cccc|c|cccc|c|c}
    \toprule
    \multicolumn{1}{c|}{\multirow{2}{*}{\textbf{Dataset}}}   & 
    \multirow{2}{*}{\textbf{Year}} &
    \multirow{2}{*}{\textbf{Region}} & 
    \multicolumn{4}{c|}{\textbf{Sensor Data}} & \multirow{2}{*}{\textbf{Frames}} &
    \multicolumn{4}{c|}{\textbf{Annotation}} &
    \multirow{2}{*}{\textbf{Weather}} &
    \multirow{2}{*}{\textbf{Time}} 
    \\
    \cline{4-7} \cline{9-12}  & & & \textbf{Scenes} & \textbf{Camera} & \textbf{LiDAR} & \textbf{Radar}  &  &  \textbf{3D Det.} & \textbf{3D Occ.} & \textbf{HD-Map} & \textbf{E2E Plan} & &   
    \\
    \midrule\midrule
    KITTI~\cite{geiger2012we} & 2012  & EU & 22 & 4$\times$  & 1$\times$64-Beam & - & 15k & \cmark & \xmark & \xmark & \xmark & \xmark & day
    \\
    ApolloScape~\cite{wang2019apolloscape}& 2018  & AS & 103  &  2$\times$ & 2$\times$VUX-1HA & -  & 144k &\cmark & \xmark  & \cmark  &  \xmark &  \cmark & day  
    \\
    nuScenes~\cite{caesar2020nuscenes} & 2019  & NA/AS & 1000   & 6$\times$ & 1$\times$32-Beam &  5$\times$3d & 40k  & \cmark  & \cmark  & \cmark     & \cmark & \cmark  & d\&n
    \\
    SemanticKITTI~\cite{behley2019iccv} & 2019  & EU & 22 & 4$\times$  & 1$\times$64-Beam & - & - & \cmark & \cmark    & \xmark &   \xmark & \xmark & day
    \\
    Waymo~\cite{sun2020scalability} & 2019 & NA & 1150 & 5$\times$ & 5$\times$64-Beam  & -    & 230k  & \cmark   & \xmark    & \xmark      & \xmark & \cmark & d\&n
    \\
    Argoverse~\cite{Argoverse} & 2019  & NA  & 113 & 7$\times$ & 2$\times$32-Beam  & -  &   22k    & \cmark  & \cmark    & \cmark     & \cmark & \cmark & d\&n
    \\
    Lyft L5~\cite{houston2020one} & 2019  & NA & 366 & 7$\times$ & 1$\times$64 \& 2$\times$40-Beam  & 5$\times$3d   & 46k   & \cmark  & \xmark    & \xmark         &   \xmark & \cmark & day
    \\
    A*3D~\cite{pham20203d} & 2019  & AS  & - & 2$\times$ &    1$\times$64-Beam   & -  &   39k    & \cmark    & \xmark     & \xmark &  \xmark & \cmark  & d\&n
    \\
    KITTI-360~\cite{Liao2021ARXIV} & 2020  &  EU     & 11 & 4$\times$  & 1$\times$64-Beam   & -  & 80k   & \cmark   & \cmark    & \xmark       &   \xmark & \xmark& day
    \\
    A2D2~\cite{geyer2020a2d2} &   2020   &   EU    & - & 6$\times$ &    5$\times$16-Beam   & -   &   12.5k     &    \cmark     & \xmark   &   \xmark      &   \xmark & \cmark & day
    \\
    PandaSet~\cite{xiao2021pandaset} & 2020  & NA & 179 & 6$\times$& 2$\times$64-Beam & - & 14k  &   \cmark   & \xmark    & \xmark     &  \xmark & \xmark  & d\&n
    \\
    Cirrus~\cite{wang2021cirrus} & 2020  & - & 12  & 1$\times$ & 2$\times$64-Beam  & -  & 6285  & \cmark  & \xmark   & \xmark     & \xmark & \xmark  & d\&n
    \\
    ONCE~\cite{mao2021one} & 2021  & AS &  -  & 7$\times$ & 1$\times$40-Beam   & - & 15k   & \cmark  & \xmark     & \xmark  &  \xmark & \cmark  & d\&n
    \\
    Shifts~\cite{malinin2021shifts} & 2021 &  AS & - &- & - & -&  - & \cmark &   \xmark   & \cmark &  \xmark &  \cmark &d\&n 
    \\
    nuPlan~\cite{caesar2021nuplan} &  2021 & NA/AS &3098  & 8$\times$ &  3$\times$40 \& 2$\times$20-Beam  & - &  -  & \cmark&\xmark &\cmark & \cmark  & \cmark& d\&n 
    \\
     Argoverse2~\cite{Argoverse2} & 2022  & NA & 1000 & 7$\times$ & 2$\times$32-Beam   & -  & 150k   & \cmark    & \cmark    & \cmark  &  \cmark & \cmark  & d\&n
    \\
    MONA~\cite{gressenbuch2022mona} &  2022 & EU &  3  & 3$\times$ &  -  & - & -  &\cmark  &\xmark      &\cmark  &\xmark   &\cmark   & day
    \\
    Dual Radar~\cite{zhang2023dual} & 2023  & AS  & 151   & 1$\times$ & 1$\times$80-Beam   & 2$\times$4d &  10k  &\cmark  &\xmark      & \xmark & \xmark  & \cmark  & d\&n
    \\
    MAN TruckScenes~\cite{fent2024man} &  2024 & EU &   747 & 4$\times$ &  6$\times$64-Beam &6$\times$4d  &  30k & \cmark &   \xmark   & \xmark & \xmark  & \cmark  &d\&n 
    \\
    OmniHD-Scenes~\cite{zheng2024omnihd} &  2024 & AS &  1501  & 6$\times$ &  1$\times$128-Beam  & 6$\times$4d & 11.9k   & \cmark& \cmark     & \cmark & \xmark  & \cmark  & d\&n
    \\
    AevaScenes~\cite{aevascenes2025}&2025&NA&100&6$\times$&6$\times$ -&-&10k&\cmark&\xmark&\cmark&\xmark&\xmark&d\&n\\
    
    PhysicalAI-AV~\cite{nvidia_physicalai_av_2025}&2025&NA/EU&310,895&7 $\times$&1$\times$ -&11$\times$&-&\xmark&\xmark&\xmark&\cmark&\cmark&d\&n\\
    \bottomrule
    \end{tabular}}
\label{tab:dataset_vehicle}
\end{table}
\section{Platform-specific Datasets} 
\label{sec:datasets}

The efficacy of multi-modal representation learning is intrinsically linked to the scale, diversity, and fidelity of the underlying data. 
As the field transitions from supervised learning to self-supervised pre-training and foundation models, the role of datasets has evolved from static benchmarks to dynamic engines for forging \textbf{Spatial Intelligence}.
In this section, we systematically evaluate prominent datasets across autonomous vehicles, aerial drones, and other robotic platforms. 
We analyze not only their sensor configurations and annotation richness but also their suitability for emerging tasks such as open-vocabulary perception and generative world modeling.

\subsection{Overview of Sensor Modalities and Datasets}

Multimodal perception systems integrate a heterogeneous suite of onboard sensors, primarily including RGB cameras, LiDAR, radar, event camera, and inertial measurement units (IMUs).
Each modality offers distinct perceptual affordances and IMUs enable high-rate ego-motion estimation.
Beyond raw sensing, the utility of a dataset for modern pre-training is defined by several critical attributes:
\begin{itemize}
  \item \textbf{Sensor Configuration \& Coverage:} The spatial arrangement and field-of-view (FoV) determine the system's ability to construct holistic 360-degree world representations.
  \item \textbf{Spatio-Temporal Synchronization:} Precise calibration is non-negotiable for learning unified representations, especially for fusing high-frequency visual streams with sparse geometric points.
  \item \textbf{Annotation Granularity \& Modality:} The shift from bounding boxes to dense occupancy grids, and recently to natural language descriptions, reflects the community's move towards reasoning-centric tasks.
  \item \textbf{Domain Diversity:} Variations in weather, lighting, and geography are essential for training robust foundation models capable of zero-shot generalization.
\end{itemize}

The following subsections examine datasets from specific platforms, revealing how platform-specific constraints shape data characteristics and subsequent learning paradigms.

\subsection{Datasets Acquired from Various Platforms}

\subsubsection{Autonomous Vehicles}
\label{sec:dataset_vehicle}

Autonomous driving serves as the primary testbed for multi-modal spatial intelligence. 
Vehicles typically deploy a redundant sensor suite consisting of surround-view cameras, high-beam LiDARs, and radars to ensure safety-critical perception~\cite{kong2023robo3d, xie2023robobev, wang2024not}.
The continuous collection of synchronized sensor streams has produced massive-scale datasets~\cite{geiger2012we, caesar2020nuscenes, sun2020scalability, nvidia_physicalai_av_2025, bari2025datasets}, which act as the fuel for self-supervised pre-training. 
Current methodologies leverage these unlabeled streams for pretext tasks such as temporal future prediction~\cite{yang2023vidar, min2024driveworld}, cross-modal masked reconstruction~\cite{lin2024bev, abdelsamad2025nomae, yang2024unipad}, and contrastive distillation~\cite{sautier2022image, xu2025beyond}, effectively turning raw data into transferable representations without human labeling.

Table~\ref{tab:dataset_vehicle} summarizes representative datasets. Notably, the evolution from early perception-centric benchmarks (\eg, KITTI~\cite{geiger2012we}) to modern reasoning-centric datasets (\eg, nuPlan~\cite{caesar2021nuplan} and Argoverse 2~\cite{Argoverse2}) highlights a crucial trend: the integration of high-definition maps, long-horizon trajectories, and increasingly, \textbf{language-based scenario descriptions}~\cite{wang2025omnidrive, liu2025occvla}. 
These rich annotations are pivotal for training next-generation End-to-End planners and Vision-Language-Action (VLA) models~\cite{li2025drivevla, survey_vla4ad, li2025end}.

\begin{table}[t] 
    \caption{\textbf{Chronological overview of state-of-the-art UAV-based datasets (2016--Present).} 
\textbf{Region}: ``Multi'' denotes data collected across multiple regions/platforms; ``Sim'' indicates synthetic simulation data. 
\textbf{Viewpoint}: ``G'' (Ground-view), ``A'' (Aerial-view), and ``AG'' (Aerial \& Ground joint view). 
\textbf{Annotations} lists the supported downstream tasks. }
    \label{tab:dataset_drone}
    \vspace{-5mm}
    \centering
    \renewcommand\tabcolsep{3pt}
    \resizebox{\textwidth}{!}{
    \begin{tabular}{r|c|c|c|c|c|c|c}
    \multicolumn{8}{c}{} 
    \\
    \midrule
    \multicolumn{1}{c|}{\multirow{1}{*}{\textbf{Dataset}}}  & \textbf{Year} & \textbf{Region} & \textbf{Viewpoint} & \textbf{Sensor Configuration} & \textbf{Frames} & \textbf{Sensor Resolution} & \textbf{Annotations} 
    \\
    \midrule\midrule
    Campus \cite{robicquet2016learning} & 2016 & NA & Single (A) & 1$\times$ Camera  & 929,499 & $1400 \times 2019$ & 
    Target Forecasting/ Tracking
    \\
    UAV123 \cite{mueller2016benchmark}  & 2016 & AS & Multi (A) & 1$\times$Camera  & 110,000 & $720 \times 720$ & UAV Tracking
    \\ %
    CarFusion\cite{reddy2018carfusion}   & 2018 & NA & Multi & 22$\times$Camera  & 53,000 & $1,280 \times 720$ & 3D Vehicle Reconstruction
    \\ 
    UAVDT \cite{du2018unmanned}   & 2018 & AS & Single & 1$\times$Camera  & 80,000 & $1080 \times 540$ & 2D object Detection/ Tracking
     \\
    DOTA\cite{xia2018dota} & 2018 & {Multi} & Single (A) & Multi-Source  & 2,806 & $4000 \times 4000$ & 2D Object Detection
    \\ 
    VisDrone\cite{zhu2021detection}   & 2019 & AS & Single (A) & 1$\times$Camera & 179,264 & $3840 \times 2160$ & 2D Object Detection/ Tracking
    \\
     DOTA V2.0~\cite{ding2021object} & 2021 & {Multi} & Single (A) & Multi-Source  & 11,268 & $4000 \times 4000$ & 2D Object Detection
     \\
     MOR-UAV \cite{mandal2020mor} & 2020 & AS & Single & 1$\times$Camera & 10,948 & $1280 \times 720$, $1920 \times 1080$ & Moving Object Recognition 
     \\
    AU-AIR \cite{bozcan2020air} & 2020 & EU & Multi & 1$\times$Camera  & 32,823 & $1920 \times 1080$ &  2D Object Detection\\
    UAVid \cite{lyu2020uavid}   & 2020 & EU & Single & 1$\times$Camera & 300 & $3840 \times 2160$, $5472 \times 3078$ & 2D Semantic Segmentation
    \\
    \multirow{2}{*}{MOHR \cite{zhang2021empirical}} & \multirow{2}{*}{2021}& \multirow{2}{*}{AS} & \multirow{2}{*}{Multi (A)} & \multirow{2}{*}{ 3$\times$Camera } & \multirow{2}{*}{10,631} & $5472 \times 43078$, $7360 \times 4192$ &\multirow{2}{*}{2D Object Detection}
    \\
     &   &   &  &    &  &   $8688 \times 5792$ &
    \\
    SensatUrban \cite{hu2021towards} & 2021 & EU & Single (A) & 1$\times$Camera & - & - & 3D Segmentation
    \\
    UAVDark135 \cite{li2022all} & 2023 & AS & Single & 1$\times$Camera & 125,466 & 1920 $\times$ 1080 & 2D Object Tracking
    \\  
    MAVREC \cite{dutta2024multiview} & 2023 & EU & Multi (AG) & 2$\times$Camera  & 537,030 & $2700 \times 1520$ & 2D Sup/Semi-Sup Object Detection
    \\ 
    BioDrone \cite{zhao2024biodrone} & 2024 & AS & Single (A) & 1$\times$Camera & 304,000 & 1400 $\times$ 1080  & 2D Object Tracking  
    \\  
    PDT \cite{zhou2025pdt} & 2024 & AS & Single (A) & 1$\times$Camera, 1$\times$LiDAR  &  5,775& 5472 $\times$ 3648, 640 $\times$ 640& 2D Object Detection
    \\  
    UAV3D \cite{ye2024uav3d} & 2024 & Sim & Multi (A) & 5$\times$Camera  & 20,000 & 800 $\times$ 450 & 3D Object Detection/ Tracking
    \\  
    IndraEye \cite{gurunath2024indraeye} & 2024 & AS & Multi (A) & 1$\times$Camera  & 2,000 & 1280 $\times$ 720, 640 $\times$ 480 &  2D Object Detection/ Semantic Segmentation
    \\  
    UAVScenes~\cite{wang2025uavscenes} & 2025 & AS & Multi (A) &1$\times$Camera, 1$\times$LiDAR  & 120,000 & 2448$\times$2048 & 2D/3D Semantic Segmentation; 6-DoF Visual Localization 
    \\
    \bottomrule
    \end{tabular}}
\vspace{0.2cm}
\end{table}

\subsubsection{Drones (UAVs)}
\label{sec:dataset_drone}

Unmanned Aerial Vehicles (UAVs) present unique perception challenges due to their bird's-eye viewpoints, six degrees-of-freedom (6-DoF) motion, and rapid scale changes~\cite{lyu2020uavid, wang2025uavscenes, hu2021towards}. 
While RGB cameras and IMUs remain standard, advanced datasets now incorporate LiDAR to capture 3D structural information for complex environments~\cite{hu2021towards, ye2024uav3d}.

Table~\ref{tab:dataset_drone} details key UAV datasets. 
Unlike ground vehicles, UAV data is characterized by significant perspective distortion and motion blur~\cite{lyu2020uavid, zhao2024biodrone}. Consequently, pre-training in this domain heavily utilizes transfer learning from ground-level or satellite imagery~\cite{zhu2021detection}, adapting visual foundation models to aerial domains. 
Recent efforts also explore cross-view geo-localization and self-supervised flow estimation to handle the dynamic nature of flight. 
The emergence of multi-modal UAV datasets~\cite{zhao2024biodrone, ye2024uav3d, wang2025uavscenes} is crucial for extending Spatial Intelligence from 2D ground planes to 3D volumetric spaces.

\begin{table}[t] 
    \caption{\textbf{Overview of multi-modal datasets for diverse robotic platforms.} This table categorizes datasets into three specialized domains: \textbf{Railways}, \textbf{Unmanned Surface Vehicles (USVs)}, and \textbf{Legged Robots}. These benchmarks extend spatial intelligence research to constrained tracks, maritime environments, and complex terrains.}
    \label{tab:dataset_others}
    \vspace{-5mm}
    \centering
    \renewcommand\tabcolsep{1.5pt}
     \resizebox{\textwidth}{!}{\begin{tabular}{r|c|c|c|c|c|c}
    \multicolumn{6}{c}{} 
    \\
    \midrule
    \multicolumn{1}{c|}{\multirow{1}{*}{\textbf{Dataset}}}  & \textbf{Year} & \textbf{Region} & \textbf{Platform} & \textbf{Sensors} & \textbf{Frames} & \textbf{Annotations} 
    \\
    \midrule\midrule
     RailSem19~\cite{zendel2019railsem19} & 2019 & EU & Railway & Camera & 8,500 & Image Classification, Semantic Segmentation \\
     FRSign~\cite{harb2020frsign} & 2020 &  EU & Railway & 2$\times$Camera  & 105,352 &  Railway Signaling Reading\\
     RAWPED~\cite{toprak2020conditional} & 2020 & EU, AS  & Railway & 1$\times$Camera  & 26,000 &  2D Object Detection \\
     SRLC~\cite{uggla2021towards} & 2021 & EU  & Railway &  LiDAR &  - & Point Cloud Generation, Semantic Segmentation  \\
     Rail-DB~\cite{li2022rail} & 2022 & AS & Railway & Camera & 7,432 & Rail Detection \\
     RailSet~\cite{zouaoui2022railset} & 2022 & EU  & Railway & 1$\times$Camera  & 6,600 & Railway Anomaly Detection \\
     OSDaR23~\cite{tagiew2023osdar23}  & 2023 & EU  &  Railway & 9$\times$Camera, 6$\times$LiDAR, 1$\times$Radar  & 1,534 & Rail and Object Detection, LiDAR Segmentation \\
     Rail3D~\cite{kharroubi2024multi} & 2024 & EU & Railway & 4$\times$Camera, 1$\times$LiDAR& - & LiDAR Semantic Segmentation \\
     WHU-Railway3D~\cite{qiu2024whu} & 2024 & AS & Railway & 1$\times$LiDAR & 40 tiles & LiDAR Segmentation \\
     FloW~\cite{cheng2021flow} & 2021 & AS& Unmanned Surface Vehicle & 2$\times$Camera, 1$\times$4D Radar &2,000 & 2D Object Detection \\ 
     DartMouth~\cite{jeong2021efficient} & 2021 & NA& Unmanned Surface Vehicle & 3$\times$Camera, 1$\times$LiDAR &- &2D Object Detection, Semantic Segmentation  \\ 
     MODS~\cite{bovcon2021mods} & 2022 & EU & Unmanned Surface Vehicle & 2$\times$Camera, 1$\times$LiDAR &8,175 &2D Object Detection  \\ 
      SeaSAW~\cite{kaur2022sea}   & 2022 & EU, NA & Unmanned Surface Vehicle & 5$\times$Camera & 1,900,000 & 2D Object Detection, Tracking, Classification \\
     WaterScenes~\cite{yao2024waterscenes}   & 2023 & AS & Unmanned Surface Vehicle & 1$\times$Camera, 1$\times$4D Radar & 54,120 & 2D Object Detection, Semantic/ Panoptic Segmentation \\
    
     MVDD13~\cite{wang2024marine} & 2024 & AS& Unmanned Surface Vehicle & Camera x1 & - & 2D  Object Detection \\ 
     SeePerSea~\cite{jeong2024multi} & 2024 & AS, NA& Unmanned Surface Vehicle & 1$\times$Camera, 1$\times$LiDAR & 10,906 & 2D \& 3D Object Detection \\ 
     WaterVG~\cite{guan2024watervg}   & 2024 & AS & Unmanned Surface Vehicle & 1$\times$Camera, 1$\times$4D Radar & 11,568 & Multi-Task Visual Grounding\\ 
     Han~\etal~~\cite{han2024lifelike} & 2024 & AS&  Legged Robots & Depth Camera & - & Animal Motions 
    
    \\
    Luo~\etal~~\cite{luo2025omnidirectional} & 2025 & AS&  Legged Robots & Panoramic Camera & 1,920s & 2D Object Tracking 

    \\
    QuadOcc~\cite{shi2025oneocc} & 2025 & AS & Legged Robots & Panoramic Camera, 1$\times$LiDAR & 8,000 & 3D Occupancy
    
    \\
    
    M3ED~\cite{chaney2023m3ed}&2023&NA&Car, UAV, Legged Robots& 3$\times$ Camera, 2$\times$ Event Camera,1$\times$LiDAR &-&Depth Estimation, Semantic Segmentation \\ 
    Pi3DET~\cite{liang2025perspective}&2025&NA&Car, UAV, Legged Robots&3$\times$ Camera, 2$\times$ Event Camera,1$\times$LiDAR &51,545&3D Object Detection \\
    \bottomrule
    \end{tabular}}
\vspace{0.2cm}
\end{table}

\subsubsection{Other Robotic Platforms}
\label{sec:dataset_others}

Beyond cars and drones, diverse robotic platforms such as Unmanned Surface Vehicles (USVs)~\cite{guan2024watervg, yao2024waterscenes}, railway systems~\cite{li2022rail, zouaoui2022railset, zendel2019railsem19}, and legged robots~\cite{han2024lifelike, chaney2023m3ed, ha2024learning} operate in highly constrained or unstructured environments. 
These domains challenge pre-training models with unique noise patterns (\eg, water reflections for USVs) and motion dynamics (\eg, non-linear locomotion for quadrupeds).

Table~\ref{tab:dataset_others} lists representative datasets. For instance, datasets for legged robots~\cite{han2024lifelike, luo2025omnidirectional} emphasize egocentric perception under severe camera shake, motivating research into robust, motion-aware representation learning. 
Similarly, rail and USV datasets focus on long-range, track-constrained perception~\cite{qiu2024whu, zouaoui2022railset, yao2024waterscenes}. 
A growing trend in these specialized domains is the use of Simulation-to-Real transfer and domain adaption~\cite{chaney2023m3ed, liang2025perspective}. 
Engines like QuaDreamer~\cite{wu2025quadreamer} generate synthetic training data to supplement scarce real-world samples, training models that can generalize to physical robots via domain randomization. 
This highlights the increasing role of synthetic data in democratizing foundation models for varied robotic form factors~\cite{uggla2021towards}.

\subsection{Key Dataset Trends and Implications}

Analyzing the landscape of platform-specific datasets identifies three evolutionary trends that are reshaping multi-modal pre-training:

\noindent \textbf{From Perception to Reasoning and Action.} 
Modern datasets are moving beyond bounding boxes. Benchmarks like nuPlan~\cite{caesar2021nuplan} and OmniDrive~\cite{wang2025omnidrive} introduce planning trajectories, logic-based scenarios, and open-vocabulary language labels. 
This shift enables the training of models that do not just \emph{see} but \emph{reason} and \emph{act}, laying the groundwork for Embodied AI and VLA models~\cite{liu2025occvla, li2025drivevla, peng2023openscene, survey_vla4ad}.

\noindent \textbf{The Rise of Synthetic and Generative Data.} 
Recognizing the long-tail limitations of real-world data, there is a surge in high-fidelity synthetic datasets and simulation environments~\cite{uggla2021towards, meier2024cdrone, wu2025quadreamer}. 
This supports the development of Generative World Models, which can simulate infinite \emph{what-if} scenarios for robust policy learning, effectively closing the loop between perception and simulation~\cite{kong20253d, worldlens}.

\noindent \textbf{Scale and Diversity for Foundation Models.} 
The explosion in data volume and modality diversity (from LiDAR to Event cameras) has rendered manual annotation obsolete~\cite{chaney2023m3ed, krispel2024maeli, yang2023event}. 
This reality firmly establishes \textbf{Self-Supervised Pre-Training} as the necessary paradigm. Future progress will depend on data engines that can automatically curate, label, and align these massive multi-modal streams to feed hungry foundation models~\cite{kong2025largead, yang2024unipad}.

These trends collectively signal a transition: datasets are no longer just static benchmarks for performance evaluation, but active components in the loop of training generative, reasoning-capable Spatial Intelligence agents~\cite{zheng2024omnihd, wang2025omnidrive, liu2025occvla}. By providing rich multi-modal contexts, these data engines facilitate the transition from passive perception to active world modeling and decision-making~\cite{kong20253d, survey_vla4ad}.
\section{Pre-Training Techniques for Perception}
\label{sec:methods}

In this section, we critically examine the methodologies that empower autonomous systems to learn robust representations from raw sensor data. 
As depicted in the taxonomy (Fig.~\ref{fig:taxonomy}), we structure the landscape based on sensor interaction paradigms: \textbf{Single-Modality} baselines, \textbf{Multi-Modality} synergy (including Camera-Centric and LiDAR-Centric distillation), and \textbf{Unified} frameworks that jointly optimize cross-modal encoders.

Beyond the modality-based categorization, we emphasize a crucial trend: the integration of \textbf{Foundation Models} and \textbf{Generative Objectives}. 
Recent approaches are shifting from simple discriminative tasks to generative reconstruction ~\cite{kerbl20233d, yang2024unipad}(\eg, NeRF, 3DGS) and future forecasting~\cite{yang2023vidar, liang2025seeing, min2024driveworld}, leveraging the rich semantic priors of large-scale vision models to enhance geometric reasoning~\cite{yu2024language, sautier2022image}. 
We also briefly discuss complementary sensors such as radar and event cameras~\cite{yao2023radar, pushkareva2024radar, yang2023event, yang2024event}. Finally, we synthesize benchmark performance to offer a holistic evaluation of how these pre-training techniques translate to downstream perception tasks.

\begin{table}[t]
    \centering
    \renewcommand\tabcolsep{3pt}
    \caption{\textbf{Comprehensive summary of LiDAR-based pre-training techniques.} The table categorizes methods into LiDAR-only (single-modality) and LiDAR-centric (cross-modal) paradigms. \textbf{Input Modality}: ``L'' denotes LiDAR input; ``SC'' and ``MC'' refer to Single-Camera and Multi-Camera data used for cross-modal distillation or alignment.}
    \vspace{-2mm}
    \resizebox{\textwidth}{!}{
    \begin{tabular}{r|r|c|c|c|c|c }
    \toprule 
    \multicolumn{1}{c|}{\multirow{1}{*}{\textbf{Method}}} & \multicolumn{1}{c|}{\multirow{1}{*}{\textbf{Venue}}} & \textbf{Input Modality} & \textbf{Proxy Task} & \textbf{Downstream Task} & \textbf{Dataset} & \textbf{Key Contribution}            
    \\ 
    \midrule\midrule
    PointContrast \cite{xie2020pointcontrast}  & ECCV'20 & L & Spatial Contra. (Point) & Sem-Seg. & ScanNet, SemanticKITTI & Point-wise contrastive learning on augmentations
    \\
    DepthContrast \cite{zhang2021self}  & ICCV'21 & L & Spatial Contra. & Sem-Seg./Det. & Waymo, nuScenes & Frame-wise depth consistency learning
    \\
    GCC-3D \cite{liang2021exploring} & {ICCV'21} & L & Spatial Contra. & Det. & Waymo & Geometry-aware contrast with clustering
    \\
    SimIPU \cite{li2022simipu} & AAAI'22 & L + SC & Spatial Contra. & Sem-Seg. & SemanticKITTI & Simple 2D-3D spatial alignment
    \\
    ProposalContrast \cite{yin2022proposalcontrast} & ECCV'22 & L & Spatial Contra. (Region) & Det. & Waymo, nuScenes & Contrastive learning on detection proposals
    \\
    GD-MAE~\cite{yang2023gd} & CVPR'23 & L & MAE & Sem-Seg./Det. & Waymo & MAE with generative decoder 
    \\
    ALSO~\cite{boulch2023also} & CVPR'23 & L & Occupancy Estimation & Occ. & nuScenes & Occupancy-based self-supervision
    \\
    BEV-MAE~\cite{lin2024bev} & AAAI'24 & L & BEV MAE & Det. & Waymo & Masked BEV feature learning 
    \\
    MAELi-MAE~\cite{krispel2024maeli} & WACV'24 & L & MAE & Det. & Waymo & MAE for large-scale LiDAR representation learning
    \\
    BEVContrast~\cite{sautier2024bevcontrast}  & 3DV'24 & L & BEV Contra. & Sem-Seg./Det. & nuScenes & Contrastive learning in BEV space
    \\
    \midrule
    PPKT~\cite{liu2021learning} & arXiv'21 & L + MC & Spatial Contra. & Sem-Seg. & nuScenes & Pixel-to-point contrastive transfer learning
    \\
    SLidR~\cite{sautier2022image} & CVPR'22 & L + MC & Spatial Contra. & Sem-Seg. & nuScenes & Superpixels to guide the image-to-LiDAR pre-training
    \\
    ST-SLidR~\cite{mahmoud2023self} & CVPR'23 & L + MC & Spatial Contra. & Sem-Seg. & nuScenes & Class-balanced cross-modal contrastive learning
    \\
    TriCC~\cite{pang2023unsupervised} & CVPR'23 & L + MC & Spatial \& Temp. Contra. & Sem-Seg. & nuScenes & Triangle-constrained spatiotemporal contrastive
    \\
    Seal~\cite{liu2024segment} & NeurIPS'23 & L + MC & Spatial Contra. & Sem-Seg. & nuScenes & Transfer knowledge from foundation models to 3D
    \\
    CSC~\cite{chen2024building} & CVPR'24 & L + MC & Spatial Distill. & Sem-Seg. & nuScenes & Unified baseline for large-scale pretraining
    \\
    OLIVINE~\cite{zhang2024fine} & NeurIPS'24 & L + MC & Spatial Distill. & Sem-Seg. & nuScenes & Fine-grained contrast with vision features
    \\
    HVDistill~\cite{zhang2024hvdistill}& IJCV'24 & L + MC & Spatial Distill. & Sem-Seg. & nuScenes & Hybrid-view distillation from images to 3D 
    \\
    ScaLR~\cite{puy2024three} & CVPR'24 & L + MC & Spatial Distill. & Sem-Seg./Det. & nuScenes, KITTI, PandaSet & Directly distill knowledge from image to LiDAR 
    \\
    SuperFlow~\cite{xu20244d} & ECCV'24 & L + MC & Spatial \& Temp. Contra. & Sem-Seg. & nuScenes & Spatiotemporal contrastive for knowledge transfer
    \\
    LargeAD~\cite{kong2025largead} & arXiv'25 & L + MC & Spatial Contra. & Sem-Seg./Det. & nuScenes, KITTI, Waymo & Large-scale multi-dataset pre-training
    \\
    LiMoE~\cite{xu2025limoe} & CVPR'25 & L + MC & Spatial \& Temp. Distill. & Sem-Seg./Det. & nuScenes & MoE-based multi-representation pre-training
    \\
    LiMA~\cite{xu2025beyond} & ICCV'25 & L + MC & Spatial \& Temp. Distill. & Sem-Seg./Det. & nuScenes & Cross-view and long-horizon distillation for pre-training
    \\
    \bottomrule
   \end{tabular}
   }
   \label{table:lidar_summary}
\vspace{0.2cm}
\end{table}

\subsection{Single-Modality Pre-Training}
\label{sec:single}

Single-modality pre-training serves as the bedrock of perception, aiming to extract intrinsic semantic and geometric features from individual sensor streams without the aid of cross-modal supervision. 
Given their ubiquity in autonomous systems, we primarily focus on \textbf{Camera} and \textbf{LiDAR} modalities in this subsection.
Mastering these single-modality representations is a prerequisite for effective sensor fusion and interaction, as it ensures that each branch of a multi-modal system contributes robust, high-quality features to the unified world model.

\subsubsection{LiDAR-Only Pre-Training}
\label{sec:lidar_only}

\begin{wrapfigure}{r}{0.5\textwidth}
    \begin{minipage}{\linewidth}
        \centering
        \vspace{-0.4cm}
        \includegraphics[width=\linewidth]{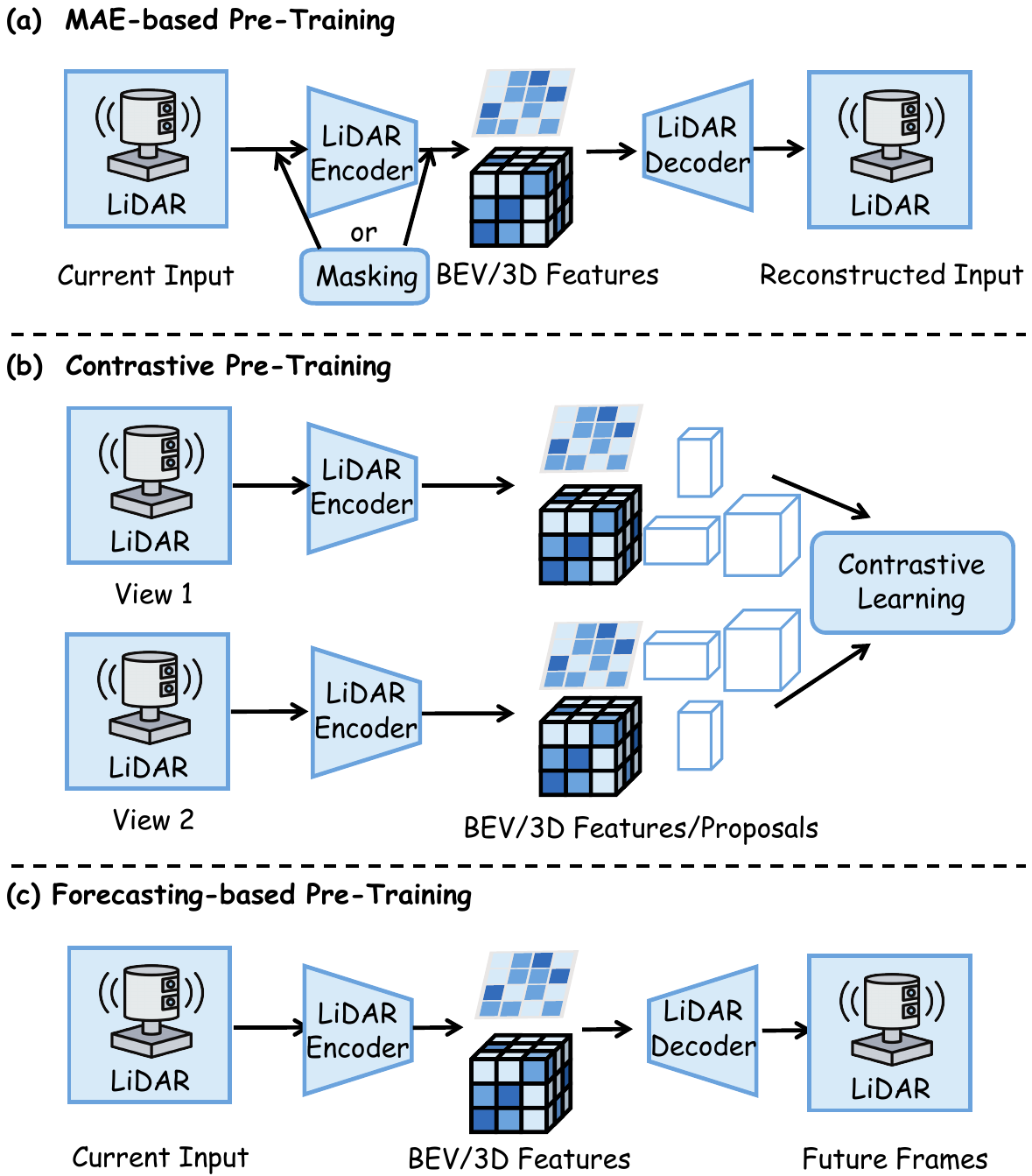}
        \vspace{-0.6cm}
        \caption{\textbf{Schematic illustration of representative LiDAR-only pre-training paradigms.} 
To learn robust geometric representations from sparse point clouds without annotations, methods typically adopt three strategies: 
\textbf{(a) Masked Autoencoding (MAE)}, which reconstructs missing structures to learn local geometry; 
\textbf{(b) Contrastive Learning}, which enforces view-invariant feature discrimination; and 
\textbf{(c) Temporal Forecasting}, which predicts future frames to capture dynamic scene evolution.}
        \label{fig:lidar_only}
    \end{minipage}
    \vspace{-0.2cm}
\end{wrapfigure}

LiDAR sensors provide precise and metric-accurate 3D measurements, making them indispensable for tasks requiring fine-grained geometric perception, such as object detection and occupancy prediction. 
Unlike cameras, LiDAR data is inherently sparse, unordered, and lacks texture, necessitating specialized pre-training objectives to capture underlying topological structures and temporal dynamics.
As illustrated in Fig.~\ref{fig:lidar_only}, current research focuses on three primary paradigms to forge robust 3D representations from unlabeled point clouds: \textbf{Masked Reconstruction} for structural understanding, \textbf{Contrastive Learning} for spatial invariance, and \textbf{Temporal Forecasting} for dynamic world modeling.

\vspace{1mm}
\noindent \textbf{Masked Reconstruction and Structural Completion.}
Drawing inspiration from Masked Autoencoders (MAE) in general vision~\cite{he2022masked} and NLP~\cite{devlin2019bert}, this paradigm forces the network to infer unseen geometric structures from partial observations, thereby learning holistic spatial priors.
To handle the irregularity of point clouds, approaches such as \textbf{GD-MAE}~\cite{yang2023gd} and \textbf{BEV-MAE}~\cite{lin2024bev} leverage regular 2D/3D grids for structured masking, while \textbf{MAELi}~\cite{krispel2024maeli} explicitly reconstructs intensity values to incorporate surface reflectivity properties. \textbf{MV-JAR}~\cite{xu2023mv} and \textbf{Occupancy-MAE}~\cite{min2023occupancy} also operate on voxelized features to enforce spatial consistency. Recent advances extend this concept to the temporal dimension. \textbf{T-MAE}~\cite{wei2024t} and \textbf{LSV-MAE}~\cite{cheng2024rethinking} reconstruct sequence-level motion patterns. Furthermore, \textbf{AD-L-JEPA}~\cite{zhu2025adljepa} moves beyond voxel reconstruction to latent space prediction, focusing on learning abstract relational reasoning rather than low-level details.

\noindent \textbf{Contrastive Learning and Spatial Invariance.}
Contrastive learning aims to learn discriminative feature spaces where semantically similar points or scenes are pulled together. This paradigm has evolved from point-level discrimination to multi-scale hierarchical understanding.
\textbf{PointContrast}~\cite{xie2020pointcontrast} pioneered this direction by optimizing point-level invariance across augmented views, while \textbf{DepthContrast}~\cite{zhang2021self} utilized single-view depth maps to construct informative pairs.
Subsequent research has scaled this objective to various spatial hierarchies: \textbf{Patch/Proposal-level} methods~\cite{nunes2022segcontrast, yin2022proposalcontrast, shrout2023patchcontrast} focus on object-centric features; \textbf{BEV-level} approaches~\cite{sautier2024bevcontrast, hegde2024equivariant, yuan2024ad} align features in the bird's-eye view for downstream perception tasks; and \textbf{Scene-level} methods~\cite{hou2021exploring, chen20224dcontrast} capture global context. 
This hierarchical evolution demonstrates the versatility of contrastive objectives in encoding geometry at different granularities.

\vspace{1mm}
\noindent \textbf{Temporal Forecasting and Predictive Modeling.}
Moving beyond static perception, forecasting-based pre-training leverages the sequential nature of LiDAR streams to anticipate future states, serving as a precursor to predictive world models.
Early works like \textbf{ALSO}~\cite{boulch2023also} and \textbf{4D-Occ}~\cite{khurana2023point} formulate pre-training as occupancy or flow prediction, enabling the model to fill in future geometric voids.
Recent frameworks such as \textbf{Copilot4D}~\cite{zhangcopilot4d} and \textbf{UnO}~\cite{agro2024uno} explicitly predict point cloud sequences, fostering temporally consistent representations.
Advanced methods further incorporate complex interactions: \textbf{PICTURE}~\cite{xu2024point} and \textbf{UNIT}~\cite{sautier2024unit} introduce mutual information maximization and spatio-temporal clustering, while \textbf{CooPre}~\cite{zhao2024coopre} and \textbf{TREND}~\cite{chen2024trend} extend forecasting to multi-agent cooperative scenarios. These approaches equip models with the predictive capacity essential for planning in dynamic environments.

\subsubsection{Camera-Only Pre-Training}
\label{sec:camera_only}

Visual data from camera offers the rich semantic information for scene understanding. 
While supervised pre-training on generic datasets like ImageNet~\cite{deng2009imagenet} and MS-COCO~\cite{lin2014microsoft} remains a standard initialization strategy for common vision backbones (\eg, ResNet~\cite{he2016resnet} and ViT~\cite{dosovitskiy2021vit}), it suffers from a domain gap when applied to the complex, 3D-centric tasks of autonomous systems. 
Consequently, the field has pivoted towards \textbf{Self-Supervised Learning (SSL)} on domain-specific onboard data, evolving through three key paradigms:

\vspace{1mm}
\noindent \textbf{Domain and Temporal Consistency.}
Handling domain shifts and exploiting temporal continuity are fundamental for robust vision. 
\textbf{INoD}~\cite{hindel2023inod} addresses the domain generalization challenge by formulating a dataset affiliation prediction pretext task, interleaving feature maps from disjoint domains to learn invariant representations. 
Capitalizing on the sequential nature of driving videos, \textbf{TempO}~\cite{lang2024self} treats region-level feature ordering as a sequence prediction problem. 
By modeling the temporal evolution of features, it enables the visual encoder to capture motion dynamics and causality, which are critical for planning.

\vspace{1mm}
\noindent \textbf{Geometric Lifting to BEV.}
Bridging the gap between 2D images and 3D perception is a core objective. 
\textbf{LetsMap}~\cite{gosala2025letsmap} pioneers a label-efficient approach for semantic Bird's-Eye-View (BEV) mapping. 
It leverages the spatial constraints inherent in monocular sequences to enforce consistency between perspective and BEV representations, effectively lifting 2D semantics into a metric space without relying on expensive dense annotations or LiDAR depth.

\vspace{1mm}
\noindent \textbf{Neural Fields and Volumetric Reasoning.}
The most recent frontier involves incorporating implicit 3D representations into visual pre-training. 
\textbf{NeRF-MAE}~\cite{irshad2025nerf} represents a paradigm shift, adapting Masked Autoencoders (MAE) to Neural Radiance Fields (NeRF). By using posed RGB images to reconstruct masked volumetric tokens, it forces the transformer to internalize 3D spatial layouts and view-dependent effects.
Similarly, \textbf{VisionPAD}~\cite{zhang2024visionpad} introduces a voxel-centric framework that combines voxel warping with multi-frame photometric consistency. This allows the model to learn fine-grained motion and geometry directly from image streams, offering a scalable alternative to depth-supervised methods.

Collectively, these methods illustrate a trajectory from learning 2D semantics to mastering 3D geometry and temporal dynamics, enabling cameras to function as standalone sensors for spatial intelligence.

\subsection{Multi-Modality Pre-Training}
\label{sec:multi_modality}

While single-modality pre-training establishes the foundational feature space, forging true Spatial Intelligence requires the synergy of heterogeneous sensors. 
The physical world manifests in diverse signals: cameras capture dense semantic texture, while LiDAR and radar provide sparse but metric-accurate geometry and kinematics. 
Multi-modality pre-training aims to bridge the \emph{semantic-geometric gap} by learning unified representations that leverage the complementary strengths of these modalities.

We categorize these approaches based on the information flow direction: \textbf{LiDAR-Centric} (distilling visual semantics into 3D geometry), \textbf{Camera-Centric} (injecting geometric priors into 2D features), and \textbf{Unified Frameworks} (jointly optimizing modality-agnostic representations). This taxonomy highlights how cross-modal interactions evolve from simple alignment to unified world modeling.

\subsubsection{LiDAR-Centric Pre-Training}
\label{sec:lidar_centric}

\begin{wrapfigure}{r}{0.5\textwidth}
    \begin{minipage}{\linewidth}
        \centering
        \includegraphics[width=\linewidth]{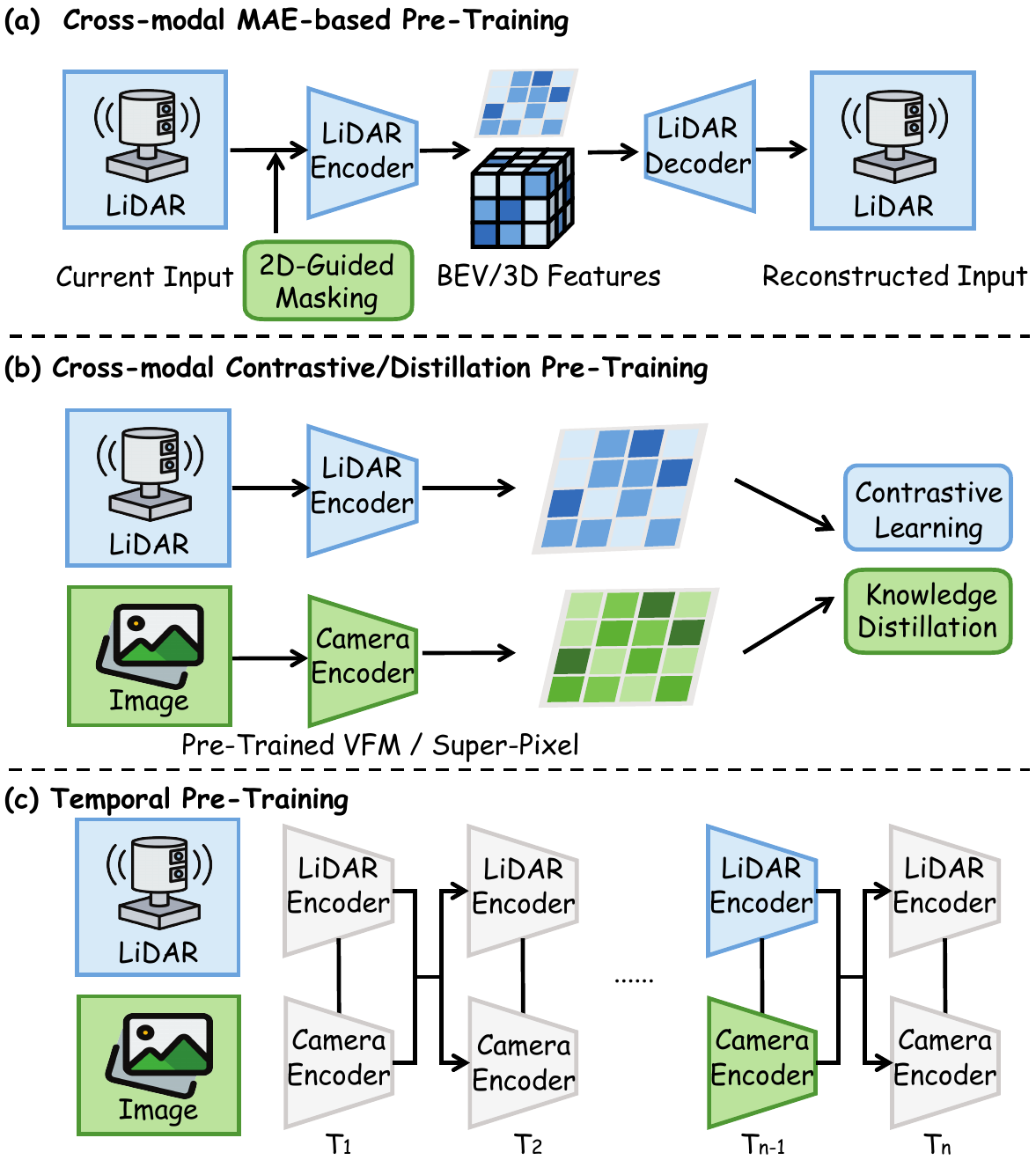}
        \vspace{-0.6cm}
        \caption{\textbf{Taxonomy of LiDAR-centric pre-training methodologies.} 
To bridge the semantic gap of point clouds, these approaches leverage images as privileged information during training. The main paradigms involve: 
\textbf{(a) Cross-modal MAE-based Pre-Training}, which incorporates \emph{2D-guided masking} strategies to enhance geometric reconstruction and structural understanding; 
\textbf{(b) Cross-modal Contrastive/Distillation Pre-Training}, which either enforces feature alignment between modalities or directly transfers rich open-vocabulary semantics from pre-trained Vision Foundation Models (VFMs) to 3D encoders; and 
\textbf{(c) Temporal Pre-Training}, which exploits video-LiDAR sequences to capture motion dynamics and enforce spatiotemporal consistency.}
        \label{fig:lidar_centric}
    \end{minipage}
    \vspace{-0.8cm}
\end{wrapfigure}

LiDAR sensors excel at capturing precise 3D structures but suffer from inherent semantic sparsity and lack of texture. 
Conversely, the computer vision community has cultivated powerful foundation models~\cite{radford2021clip, kirillov2023sam, caron2021dino} that encapsulate rich, open-world semantic knowledge.
LiDAR-centric pre-training aims to bridge this asymmetry by treating visual signals as \emph{Privileged Information} during training. 
The goal is to transfer the semantic richness of 2D images into 3D point cloud networks, enabling them to hallucinate semantic features even when cameras are absent during inference. As illustrated in Fig.~\ref{fig:lidar_centric}, this paradigm has evolved through four key strategies:

\vspace{1mm}

\noindent \textbf{Masked Reconstruction with Visual Guidance.}
Integrating cross-modal cues from camera images into the Masked Autoencoder (MAE) framework~\cite{he2022masked} enhances structural learning for LiDAR point cloud.
\textbf{I2P-MAE}~\cite{zhang2023learning} and \textbf{CM3D}~\cite{khuranashelf} condition the reconstruction of masked LiDAR tokens on visible image patches, forcing the network to infer 3D geometry from 2D semantic context.
\textbf{ImageTo360}~\cite{reichardt2023360} and \textbf{EUCA-3DP}~\cite{sun2024exploring} extend this to full-scene scales, leveraging BEV context to promote holistic spatial reasoning that fuses visual texture with geometric occupancy.

\noindent \textbf{Cross-Modal Contrastive Alignment.}
The foundational approach involves aligning 2D and 3D feature spaces through contrastive learning. 
By maximizing the similarity between corresponding image pixels and projected LiDAR points, models learn to associate geometric clusters with visual concepts.
\textbf{SimIPU}~\cite{li2022simipu} and \textbf{SLidR}~\cite{sautier2022image} pioneered this by constructing point-pixel pairs to enforce local semantic consistency. 
Recent extensions like \textbf{ST-SLidR}~\cite{mahmoud2023self} and \textbf{Cross-Modal SSL}~\cite{cai2024cross} incorporate temporal constraints and region-aware affinity, improving the robustness of alignment against calibration errors and dynamic objects.

\vspace{1mm}

\noindent \textbf{Knowledge Distillation from Foundation Models.}
Moving beyond simple alignment, recent works leverage 2D foundation models as \emph{teachers} to distill open-vocabulary semantics into 3D \emph{students}.
\textbf{Seal}~\cite{liu2024segment} and \textbf{ScaLR}~\cite{puy2024three} utilize the segmentation capability of SAM and vision transformers to generate high-quality pseudo-labels or soft feature targets for point clouds.
\textbf{CSC}~\cite{chen2024building} and \textbf{OLIVINE}~\cite{zhang2024fine} further refine this process by incorporating hierarchical clustering and class-aware gating, ensuring that the distilled knowledge respects the geometric boundaries of 3D objects. 
This strategy effectively imparts sight to blind LiDAR networks.

\vspace{1mm}
\noindent \textbf{Temporal Dynamics and Motion Transfer.}
Static cross-modal alignment is insufficient for dynamic autonomous systems in real world. 
\textbf{SuperFlow}~\cite{xu20244d} and \textbf{PRED}~\cite{yang2024pred} introduce temporal supervision by transferring motion knowledge from video to point cloud sequences. By aligning the temporal evolution of features across modalities, these methods enable LiDAR backbones to capture long-horizon dynamics~\cite{xu2025beyond}, serving as a stepping stone towards predictive world models.

In summary, LiDAR-centric pre-training transforms point cloud networks from pure geometric processors into semantically aware perception engines, significantly enhancing performance in detection and semantic segmentation tasks, particularly in data-scarce regimes.
Table~\ref{table:lidar_summary} provides a comprehensive taxonomy of these LiDAR-based techniques, categorizing them by input modality, proxy tasks, and downstream applications.

\begin{table}[t]
   \centering
   \renewcommand\tabcolsep{3pt}
   \caption{\textbf{Overview of camera-centric and unified pre-training methodologies.} This table summarizes representative approaches that leverage visual data as the primary input. \textbf{Input Modality}: ``MC'' denotes Multi-Camera setups; ``L'' indicates the use of LiDAR; ``T'' signifies the integration of temporal information for dynamic modeling.}
   \vspace{-2mm}
   \resizebox{\textwidth}{!}{
   \begin{tabular}{r|r|c|c|c|c|c }
   \toprule 
   \multicolumn{1}{c|}{\multirow{1}{*}{\textbf{Method}}} & \multicolumn{1}{c|}{\multirow{1}{*}{\textbf{Venue}}}  & \textbf{Input Modality} & \textbf{Proxy Task}        & \textbf{Downstream Task} & \textbf{Dataset} & \textbf{Key Contribution}            
   \\ 
   \midrule\midrule
   GeoMIM~\cite{liu2023geomim}  & {ICCV'23} & MC & Reconstruction & Det./Map/Occ. & nuScene & Leveraging the knowledge of a pretrained LiDAR model \\
   OccNet~\cite{tong2023scene}  & {ICCV'23} & MC & Forecasting & Det./Map & nuScene & Utilizing the semantic occupancy as the latent feature supervision
   \\
   UniScene~\cite{min2024multi}  & {RA-L'24} & MC & Forecasting & Det./Occ. & nuScene & Utilizing the geometric occupancy as the latent feature supervision
   \\
   DriveWorld~\cite{min2024driveworld}  & {CVPR'24} & MC & Forecasting & Det./Map/Occ./E2E & nuScene & Utilizing 4D occupancy as the latent feature supervision
   \\
   ViDAR~\cite{yang2023vidar}  & {CVPR'24} & MC-T & Forecasting & Det./Map/Occ./E2E & nuScene & Visual point cloud forecasting \\
   MIM4D~\cite{zou2024mim4d}  & {IJCV'25} & MC-T & Rendering & Det./Map/Vec. Map & nuScene &  Investigating spatial and temporal relations with video \\
   GaussianPretrain~\cite{xu2024gaussianpretrain}  & {arXiv'24} & MC-T & Rendering & Det./Occ./Vec. Map & nuScene & Leveraging the Gaussian representation \\
   VisionPAD~\cite{zhang2024visionpad}  & {CVPR'25} & MC-T & Rendering & Det./Occ.  & nuScene & Vision-only pre-training with temporal constraint \\
   \midrule
   UniPAD~\cite{yang2024unipad}  & {CVPR'24} & MC \& L & Rendering & Det./Seg. & nuScene & Multi-modality pre-training with MAE \\
   UniM2AE~\cite{zou2024unim}  & {ECCV'24} & MC \& L & Rendering &  Det./Map & nuScene & Multi-modality pre-training with MAE and extra alignment \\
    NS-MAE~\cite{xu2024learningshared}  & {CASE'25} & MC \& L & Rendering & Det./Map & nuScene &  Multi-modality pre-training with differential
neural volume rendering \\
    BEVWorld~\cite{zhang2024bevworld}  & {arXiv'24} & MC-T \& L & Rendering & Det./Motion & nuScene & Multi-modality with temporal information \\
    LRS4Fusion~\cite{palladin2025self} & ICCV'25 &  MC-T \& L & Forecasting &  Det./Depth  &   LR \& nuScenes  & Self-supervised sparse sensor fusion for long range perception
    \\
   
    \bottomrule
   \end{tabular}
}
\label{table:vision_summary}
\end{table}
\subsubsection{Camera-Centric Pre-Training}
\label{sec:camera_centric}

\begin{wrapfigure}{r}{0.55\textwidth}
    \begin{minipage}{\linewidth}
        \centering
        \includegraphics[width=\linewidth]{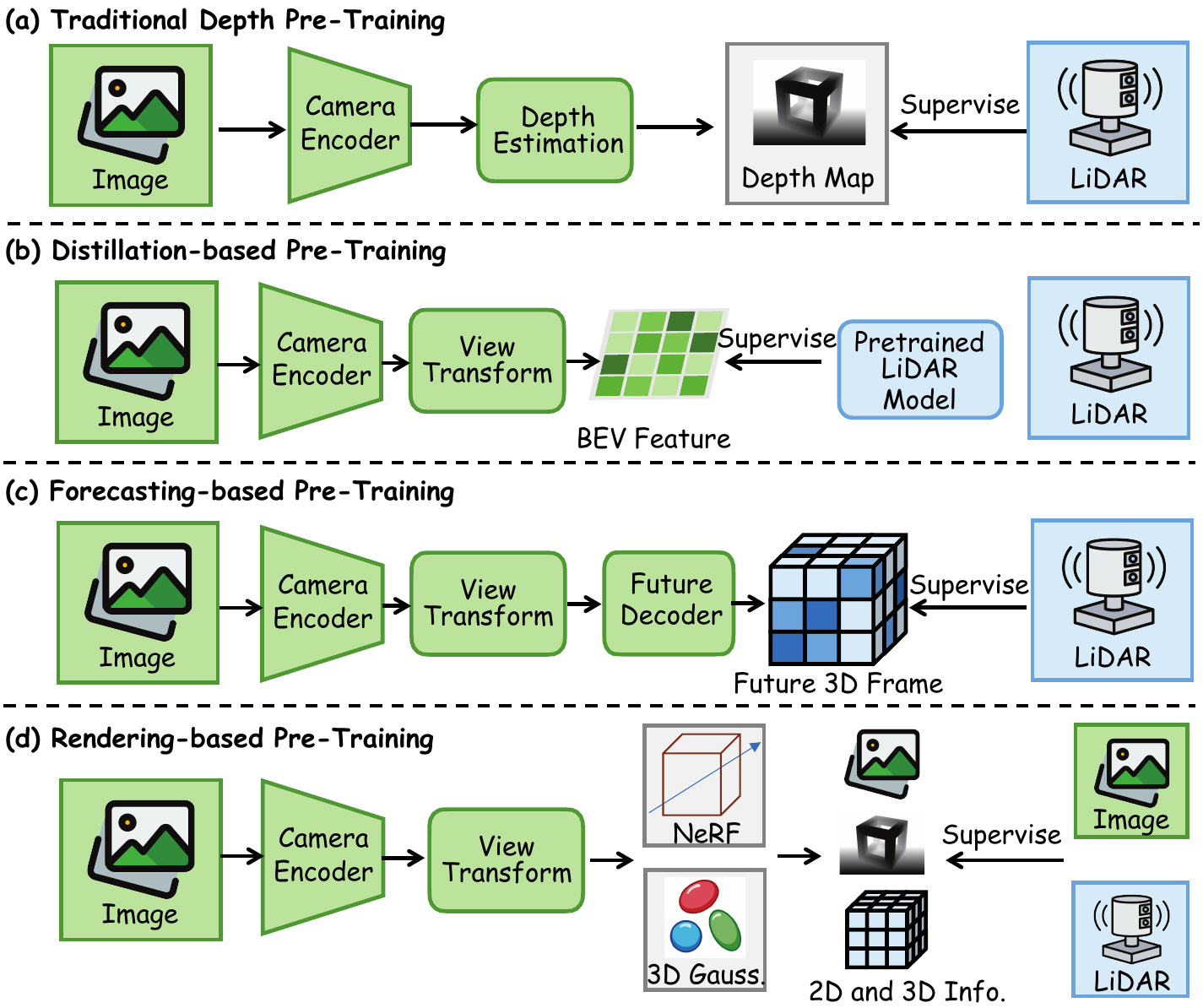}
        \vspace{-0.6cm}
        \caption{\textbf{Overview of camera-centric pre-training paradigms (LiDAR-to-Vision).} 
These methods aim to inject 3D geometric priors into 2D visual backbones using LiDAR as a supervisor. Key approaches include: 
(a) \textbf{Depth Estimation} for explicit geometry learning; 
(b) \textbf{Feature Distillation} to align 2D-3D latent spaces; and 
(c) \textbf{Forecasting}  and (d) \textbf{Generative Rendering}, which empower vision models to hallucinate 3D structures and predict future dynamics from monocular inputs.}
         \label{fig:vision_centric}
    \end{minipage}
    \vspace{-0.4cm}
\end{wrapfigure}

Camera-centric pre-training addresses the ill-posed nature of monocular perception: recovering 3D structures from 2D projections. 
While cameras are cost-effective and ubiquitous, they lack intrinsic and accurate depth. 
To overcome this, recent methods utilize LiDAR data as a \emph{Geometric Supervisor} during pre-training. 
By injecting precise depth and structural priors into visual backbones, these models learn to hallucinate 3D geometry from images alone, retaining efficient camera-only inference while benefiting from LiDAR-grade supervision. 
As visually taxonomized in Fig.~\ref{fig:vision_centric}, this domain bifurcates into two primary streams: \emph{Geometric Perception} (via explicit depth or feature distillation) and \emph{Predictive World Modeling} (via forecasting or neural rendering).
A detailed overview of these vision-centric methodologies, including their proxy tasks and key contributions, is summarized in Table~\ref{table:vision_summary}.


\noindent \textbf{Geometric Perception: From Depth to Distillation.}
The primary goal here is to equip vision models with spatial awareness by aligning 2D features with 3D structural constraints.
\emph{Explicit Depth Pre-Training} (Fig.~\ref{fig:vision_centric} (a)) serves as the foundational approach. 
Early works like \textbf{DD3D}~\cite{park2021pseudo} and \textbf{DEPT}~\cite{li2022delving} leverage pseudo-depth supervision from LiDAR to initialize 3D object detectors, effectively grounding visual features in metric space.
Moving beyond simple depth maps, \emph{Distillation-based Pre-Training} (Fig.~\ref{fig:vision_centric} (b)) aligns latent representations.
\textbf{OccNet}~\cite{tong2023scene} and \textbf{SelfOcc}~\cite{huang2024selfocc} advance this by learning to predict dense 3D occupancy grids, utilizing LiDAR occupancy as a ground-truth supervisor.
Furthermore, Masked Image Modeling (MIM) has been adapted for geometric consistency: \textbf{GeoMIM}~\cite{liu2023geomim} and \textbf{MIM4D}~\cite{zou2024mim4d} reconstruct masked image patches by cross-referencing with projected LiDAR points, forcing the network to internalize 3D spatial correspondences within the feature extraction process.

\vspace{1mm}
\noindent \textbf{Predictive World Modeling: Forecasting and Rendering.}
This stream represents the transition from static perception to dynamic simulation, requiring models to understand temporal evolution and photorealistic synthesis.
\emph{Forecasting-based Pre-Training} (Fig.~\ref{fig:vision_centric} (c)) compels models to predict future states from current video streams, thereby internalizing the physics of the environment.
\textbf{ViDAR}~\cite{yang2023vidar} pioneers "Visual Point Cloud Forecasting," treating future LiDAR points as a supervision signal for historical visual inputs.
Extensions like \textbf{DriveWorld}~\cite{min2024driveworld} and \textbf{UniScene}~\cite{min2024multi} scale this to 4D occupancy, learning spatio-temporal abstractions that facilitate long-term planning.
Complementing this, \emph{Rendering-based Pre-Training} (Fig.~\ref{fig:vision_centric} (d)) exploits the differentiability of neural fields.
Frontier methods like \textbf{GaussianPretrain}~\cite{xu2024gaussianpretrain} and \textbf{GaussianOcc}~\cite{gan2024gaussianocc} incorporate 3D Gaussian Splatting (3DGS)~\cite{kerbl20233d} into the pre-training loop. 
By enforcing photometric consistency through differentiable rendering, these models learn continuous, high-fidelity geometric representations that surpass discrete voxels in precision.
Finally, generative approaches such as \textbf{GenAD}~\cite{yang2024generalized} and \textbf{OccSora}~\cite{wang2024occsora} integrate these concepts to function as neural simulators, paving the way for end-to-end agents capable of reasoning about future consequences~\cite{yang2024driving, wei2024occllama}.

\subsubsection{Unified Pre-Training}
\label{sec:unified}

Unified pre-training represents the convergence of multi-modal learning. 
Unlike asymmetric distillation (LiDAR-centric or Camera-centric), which treats one modality as primary, unified frameworks jointly optimize encoders for heterogeneous modalities within a shared latent space. 
As explicitly illustrated in Fig.~\ref{fig:unified}, a canonical unified framework processes data through a cohesive pipeline encompassing masking, alignment, and reconstruction. This paradigm can be deconstructed into three critical phases:

\vspace{1mm}
 \begin{wrapfigure}{r}{0.66\textwidth}
    \begin{minipage}{\linewidth}
        \centering
        \vspace{-0.4cm}
        \includegraphics[width=\linewidth]{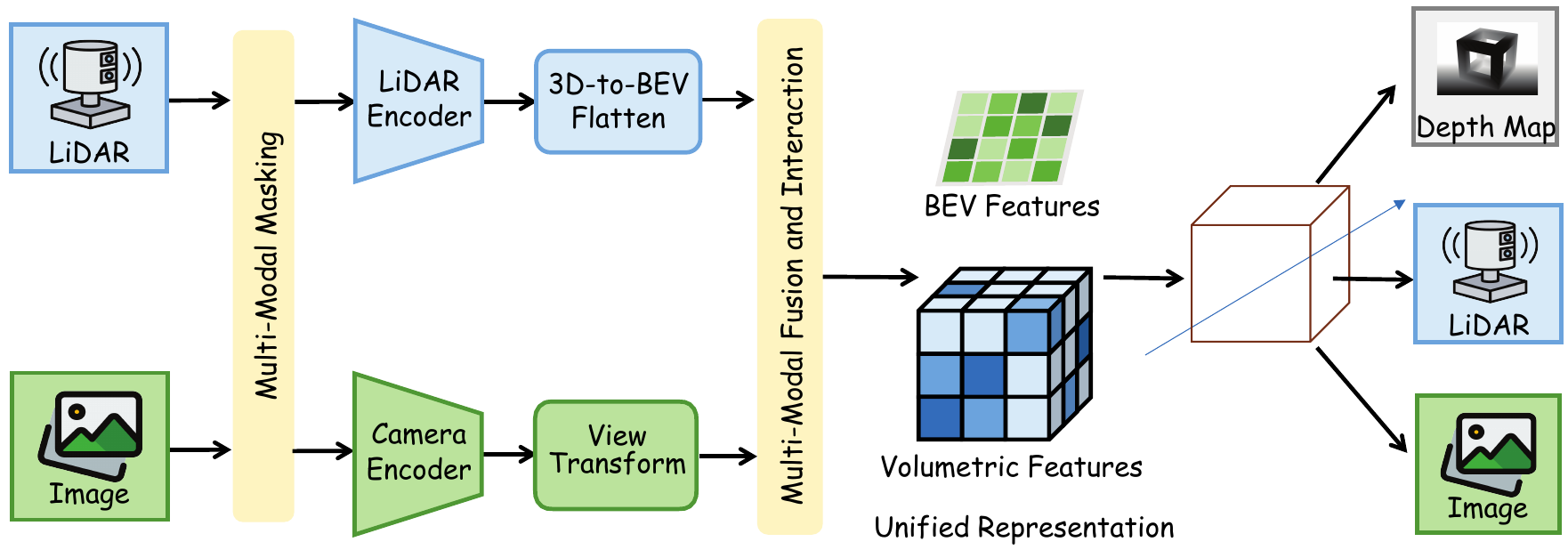}
        \vspace{-0.6cm}
        \caption{\textbf{Illustration of unified multi-modal pre-training frameworks.} 
Unlike asymmetric distillation, unified approaches jointly optimize Camera and LiDAR encoders within a shared representation space. 
This paradigm facilitates the learning of \emph{modality-agnostic} features that integrate both semantic richness and geometric precision, forging a holistic basis for Spatial Intelligence.}
        \label{fig:unified}
    \end{minipage}
    \vspace{-0.2cm}
\end{wrapfigure}
\noindent \textbf{Multi-Modal Masking and Encoding.}
The pipeline begins by treating raw sensor inputs as discrete tokens. As depicted in the \emph{Multi-Modal Masking} stage of Fig.~\ref{fig:unified}, methods like \textbf{UniPAD}~\cite{yang2024unipad} and \textbf{UniM2AE}~\cite{zou2024unim} apply randomized masking to both LiDAR points and image patches. 
This forces the encoders to learn robust local features rather than relying on redundant shortcuts. 
Specifically, the visual branch typically employs a \emph{Camera Encoder}, while the geometric branch utilizes a \emph{LiDAR Encoder} to extract high-dimensional primitives from sparse inputs.

\vspace{1mm}
\noindent \textbf{View Transformation and Unified Fusion.}
To bridge the dimensional gap between 2D images and 3D points, the framework transforms heterogeneous features into a common coordinate system. As shown in the center of Fig.~\ref{fig:unified}, visual features undergo a \emph{View Transform}~\cite{li2022bevformer, yang2024unipad, li2023delving}, while point features are processed via \emph{3D-to-BEV Flattening}. These streams converge at the \emph{Multi-Modal Fusion and Interaction} stage, resulting in a \textbf{Unified Representation} -- manifesting typically as \emph{BEV Features} or dense \emph{Volumetric Features}. Approaches like \textbf{BEVWorld}~\cite{zhang2024bevworld} and \textbf{GS3}~\cite{liu2024point} leverage this shared latent space to enforce strict geometric consistency between modalities.

\vspace{1mm}
\noindent \textbf{Generative Reconstruction.}
The final objective is to validate the understanding of the scene by reconstructing the masked or missing information. The right side of Fig.~\ref{fig:unified} demonstrates that the unified representation is decoded to simultaneously reconstruct the original \emph{LiDAR} geometry, \emph{Image} texture, and often auxiliary \emph{Depth Maps}. By optimizing for this holistic reconstruction objective, the model learns modality-agnostic features that integrate semantic richness with geometric precision, ensuring robustness even when individual sensors are compromised during inference~\cite{chen2024clap, zhang2024condense, xu2024learningshared}.

In conclusion, unified pre-training moves beyond simple sensor fusion; it forges a holistic understanding of the physical world that is independent of the specific sensing apparatus, a key characteristic of true Spatial Intelligence.

\subsection{Incorporating Additional Sensors}
\label{sec:other_sensors}

In complex open-world environments, reliance solely on cameras and LiDAR can lead to perceptual failures under adverse conditions, such as severe weather, high-speed motion, or extreme lighting changes~\cite{kong2023robo3d, xie2023robobev, kong2023robodepth, kong2025eventfly}. 
To forge robust Spatial Intelligence, incorporating complementary sensors becomes imperative.
Millimeter-wave radar offers resilience against fog and rain via Doppler signatures, while Event Cameras (neuromorphic sensors) capture microsecond-level dynamics with high dynamic range.
Integrating these modalities into pre-training frameworks not only enhances system reliability but also extends the operational design domain of autonomous agents. 
In this subsection, we analyze representation learning paradigms specialized for these sensors.

\subsubsection{Radar Pre-Training}
Radar point clouds differ significantly from LiDAR in their scarcity, noise characteristics (clutter), and unique velocity channels. Pre-training methods in this domain focus on suppressing noise and extracting meaningful structural features through three key approaches:

\vspace{1mm}
\noindent \textbf{Cross-Modal Alignment and Supervision.}
Due to the semantic sparsity of radar returns, aligning them with richer modalities is a standard strategy.
\textbf{AssociationNet}~\cite{dong2021radar} utilizes well-structured LiDAR point clouds to supervise radar feature learning, enhancing geometric consistency.
\textbf{RadarContrast}~\cite{wang2024self} and \textbf{RiCL}~\cite{decourt2024leveraging} employ contrastive learning to enforce invariance between radar representations and their multi-view or temporal counterparts, effectively grounding radar features in a stable metric space.

\vspace{1mm}
\noindent \textbf{Masked Modeling for Sparse Signals.}
Adapting masked reconstruction to radar involves dealing with extreme sparsity.
\textbf{MVRAE}~\cite{zhu2024multi} and \textbf{RSLM}~\cite{pushkareva2024radar} introduce autoencoding frameworks that reconstruct raw radar signals, enabling the model to learn spatiotemporal priors and filter out multi-path noise.
\textbf{Radar-Rep}~\cite{yao2023radar} and \textbf{Radical}~\cite{hao2024bootstrapping} further refine this by designing radar-specific masking strategies and curriculum learning to handle the high variance in signal quality.

\vspace{1mm}
\noindent \textbf{Domain Adaptation and Simulation.}
To bridge the gap between synthetic and real-world radar data, domain-adaptive strategies are crucial. \textbf{SS-RODNet}~\cite{zhuang2023effective, zhuang2023pre} facilitates transfer learning across domains, while \textbf{U-MLPNet}~\cite{yan2024learning} explores lightweight inductive biases to enable efficient radar perception on edge devices.

\subsubsection{Event Camera Pre-Training}
Event cameras capture asynchronous brightness changes, offering a paradigm shift for high-speed perception. 
Pre-training methodologies here must address the non-grid, asynchronous nature of event streams:

\vspace{1mm}
\noindent \textbf{Spatiotemporal Reconstruction.}
Reconstructing dense signals from sparse events forces the model to understand scene dynamics.
\textbf{MEM}~\cite{klenk2024masked} and \textbf{DMM}~\cite{huang2024data} adapt masked modeling to event streams, reconstructing spatial structures from fragmented temporal triggers.
\textbf{ECDP}~\cite{yang2023event} and \textbf{ECDDP}~\cite{yang2024event} focus on future frame prediction, leveraging the high temporal resolution of events to forecast motion with exceptional precision.
\textbf{STP}~\cite{liangenhancing} introduces specialized transformer architectures to simultaneously model the spatial sparsity and temporal continuity inherent in event data.

\vspace{1mm}
\noindent \textbf{Cross-Modal Synergy.}
Integrating events with standard RGB frames combines high dynamic range with semantic texture.
\textbf{EventBind}~\cite{zhou2024eventbind} aligns asynchronous event streams with synchronous RGB frames in a shared latent space, enabling semantic understanding even in high-motion blur scenarios.
\textbf{EventFly}~\cite{kong2025eventfly} further demonstrates the utility of this synergy for agile navigation in aerial robotics, where latency is a critical bottleneck.

\subsubsection{Auxiliary Modalities}
Beyond primary perception sensors, other onboard instruments serve as critical sources of \emph{weak supervision} or \emph{geometric constraints} during pre-training~\cite{geiger2012we, caesar2020nuscenes, Argoverse}, rather than just as input modalities:

\begin{itemize}
    \item \textbf{Inertial Measurement Units (IMU):} Instead of learning IMU representations in isolation, recent works utilize IMU data to enforce ego-motion consistency. By providing accurate acceleration and orientation priors, IMUs supervise the temporal alignment of vision and LiDAR backbones, essential for learning physically plausible world models.
    \item \textbf{GPS and Localization Signals:} Global positioning data provides coarse-grained location context. In large-scale pre-training, GPS traces are often used to retrieve topologically neighboring scenes or to enforce trajectory consistency in long-horizon prediction tasks.
    \item \textbf{Thermal/Infrared Sensors:} In safety-critical applications, these sensors provide distinct signatures for living beings (\eg, pedestrians and animals) that are invisible to standard cameras at night. Pre-training on thermal data typically follows domain adaptation paradigms to transfer RGB-based semantic knowledge to the thermal domain.
\end{itemize}

\begin{table}
    \centering
    \renewcommand\tabcolsep{3pt}
    \caption{\textbf{Comparative analysis of 3D object detection on the nuScenes benchmark~\cite{caesar2020nuscenes}.}  The table reports the mean Average Precision (mAP) and NuScenes Detection Score (NDS) of various pre-training frameworks. The values in parentheses denote the performance gains relative to the corresponding baseline methods.}
    \vspace{-2mm}
    \resizebox{\textwidth}{!}{
    \begin{tabular}{r|r|c|c|c|ccc|ccc}
    \toprule
    \multicolumn{1}{c|}{\multirow{2}{*}{\textbf{Method}}} & \multicolumn{1}{c|}{\multirow{2}{*}{\textbf{Venue}}} & \multirow{2}{*}{\textbf{Backbone}} & \multirow{2}{*}{\textbf{Image Size}}  & \multirow{2}{*}{\textbf{Baseline}} & \multicolumn{3}{c}{\textbf{Auxiliary Data}} & \multicolumn{2}{|c}{\textbf{Performance}}
    \\
    ~ & ~ & ~ & ~ & ~ & Temporal & Pre-Training & Others & mAP & NDS  
    \\
    \midrule\midrule
     FCOS3D~\cite{wang2021fcos3d} & {ICCVW'21} & ResNet101~\cite{he2016resnet} & $1600 \times 900$ & BEVFormer~\cite{li2022bevformer} & \cmark & ImageNet~\cite{deng2009imagenet} & - & $41.6$\dplus{$+3.9$} & $51.7$ \dplus{$+4.0$}   
    \\
    GeoMIM~\cite{liu2023geomim} & {ICCV'23} & Swin-B~\cite{liu2021swin} & $1408 \times 512$ & BEVDepth~\cite{li2023bevdepth} & \xmark & ImageNet~\cite{deng2009imagenet} & - & $52.3$\dplus{$+5.7$} & $60.5$ \dplus{$+5.0$}   
    \\
    OccNet~\cite{tong2023scene} & {ICCV'23} & ResNet101~\cite{he2016resnet} & $1600 \times 900$ & BEVFormer~\cite{li2022bevformer} & \cmark &
    ImageNet~\cite{deng2009imagenet} & - & $43.6$\dplus{$+2.0$} & $53.2$ \dplus{$+1.5$}   
    \\
     UniScene~\cite{min2024multi} & {RA-L'24} & ResNet101~\cite{he2016resnet} & $1600 \times 900$ & BEVFormer~\cite{li2022bevformer} & \cmark & FCOS3D~\cite{wang2021fcos3d} & - & $43.8$\dplus{$+2.2$} & $53.4$ \dplus{$+1.7$}   
    \\
    DriveWorld~\cite{min2024driveworld} & {CVPR'24} & ResNet101~\cite{he2016resnet} & $1600 \times 900$ & BEVFormer~\cite{li2022bevformer} & \cmark &
    FCOS3D~\cite{wang2021fcos3d} & - & $44.2$\dplus{$+2.6$} & $53.6$ \dplus{$+1.9$}   
    \\
    ViDAR~\cite{yang2023vidar} & {CVPR'24} & ResNet101~\cite{he2016resnet} & $1600 \times 900$ & BEVFormer~\cite{li2022bevformer} & \cmark & FCOS3D~\cite{wang2021fcos3d} & - & $45.8$\dplus{$+4.3$} & $54.8$ \dplus{$+4.3$}   
    \\
    UniPAD~\cite{yang2024unipad} & {CVPR'24} & ConNeXt-S~\cite{liu2022convnet} & $1600 \times 900$ & UVTR-CS~\cite{li2022uvtr} & \xmark & FCOS3D~\cite{wang2021fcos3d} & - & $42.8$\dplus{$+3.6$} & $50.2$ \dplus{$+1.4$}   
    \\
    MIM4D~\cite{zou2024mim4d} & {IJCV'25} & ResNet50~\cite{he2016resnet} & ~$704 \times 256$ & Sparse4Dv3~\cite{lin2023sparse4d} & \cmark & ImageNet~\cite{deng2009imagenet} & - & $46.4$\dplus{$+0.1$} & $57.0$ \dplus{$+0.6$}   
    \\
    GaussianPretrain~\cite{xu2024gaussianpretrain} & {arXiv'24} & ResNet50~\cite{he2016resnet} & $1600 \times 900$ & StreamPETR~\cite{wang2023streampetr} & \xmark & ImageNet~\cite{deng2009imagenet} & - & $38.6$\dplus{$+0.6$} & $48.8$ \dplus{$+0.9$}   
    \\
    VisionPAD~\cite{zhang2024visionpad} & {CVPR'25} & ResNet101~\cite{he2016resnet} & $1600 \times 900$ & UVTR-CS~\cite{li2022uvtr} & \cmark & FCOS3D~\cite{wang2021fcos3d} & - & $43.1$\dplus{$+3.9$} & $50.4$ \dplus{$+1.6$}   
    \\
    SQS~\cite{zhang2025sqs} & {NeurIPS'25} & ResNet101~\cite{he2016resnet} & $1408 \times 512$ & SparseBEV~\cite{liu2023sparsebev} & \cmark & FCOS3D~\cite{wang2021fcos3d} & - & $50.9$\dplus{$+0.8$} & $60.2$ \dplus{$+1.0$}   
    \\
    \midrule
    UniPAD~\cite{yang2024unipad} & {CVPR'24} & VoxelNet~\cite{zhou2018voxelnet} & - & UVTR-L~\cite{li2022uvtr} & \xmark & - & - & $65.0$\dplus{$+4.1$} & $70.6$ \dplus{$+2.9$}   
    \\
    UniM2AE~\cite{zou2024unim} & {ECCV'24} & SST~\cite{fan2022embracing} & - & TransFusion~\cite{bai2022transfusion} & \xmark & - & - & $65.7$\dplus{$+0.7$} & $70.4$ \dplus{$+0.5$}   
    \\
    \midrule
    NS-MAE~\cite{xu2024learningshared} & {arXiv'24} & VoxelNet~\cite{zhou2018voxelnet}+Swin-T~\cite{liu2021swin} & ~$704 \times 256$ & BEVFusion~\cite{liu2023bevfusion} & \xmark & - & - & $63.0$\dplus{$+2.2$} & $65.5$ \dplus{$+1.4$}   
    \\
    UniPAD~\cite{yang2024unipad} & {CVPR'24} & VoxelNet~\cite{zhou2018voxelnet}+ConNeXt-S~\cite{liu2022convnet} & $1600 \times 900$ & UVTR-M~\cite{li2022uvtr} & \xmark & FCOS3D~\cite{wang2021fcos3d} & - & $69.9$\dplus{$+4.5$} & $73.2$ \dplus{$+3.0$}   
    \\
    UniM2AE~\cite{zou2024unim} & {ECCV'24} & SST~\cite{fan2022embracing}+Swin-T~\cite{liu2021swin} & $1600 \times 900$ & FocalFormer3D~\cite{chen2023focalformer3d} & \xmark & MMIM~\cite{zou2024unim} & - & $71.1$\dplus{$+0.6$} & $73.8$ \dplus{$+0.7$}   
    \\
    \bottomrule
\end{tabular}}
\label{table:nus_det}
\end{table}

\subsection{Empirical Analysis and Benchmark Performance}
\label{sec:benchmark}

To empirically substantiate the efficacy of the discussed pre-training paradigms, we evaluate their impact on core 3D perception tasks: 3D Object Detection and LiDAR Semantic Segmentation. 
These tasks serve as the definitive litmus test for \textbf{Spatial Intelligence}, assessing whether learned representations can translate pretext objectives (\eg, reconstruction and forecasting) into precise geometric localization and fine-grained semantic understanding.
In this subsection, we synthesize key findings from major benchmarks, highlighting how different pre-training strategies reshape the performance landscape.

\subsubsection{3D Object Detection}
3D object detection requires the model to identify and localize objects within a metric space, a task that demands both high-level semantics and low-level geometric precision.
Quantitative results on the nuScenes benchmark (Table~\ref{table:nus_det}) provide compelling evidence for the superiority of \textbf{Unified Pre-Training}.

As shown in the comparative analysis, frameworks that jointly optimize multi-modal encoders consistently outperform camera-only baselines. 
Notably, \textbf{UniM2AE}~\cite{zou2024unim} achieves state-of-the-art performance with $71.1$ mAP and $73.8$ NDS, representing a significant gain over the strong FocalFormer3D~\cite{chen2023focalformer3d} baseline.
Similarly, \textbf{UniPAD}~\cite{yang2024unipad} demonstrates remarkable robustness, boosting the UVTR-M~\cite{li2022uvtr} baseline by +$4.5$ mAP to reach $69.9$ mAP.
This suggests that learning a shared latent space for vision and geometry allows the model to capture complementary features that are otherwise lost in late-fusion pipelines, proving that unified multi-modal masking is superior to disjoint training strategies.

\begin{table}
    \centering
    \renewcommand\tabcolsep{3pt}
    \caption{\textbf{Benchmark of cross-modal pre-training for LiDAR semantic segmentation on nuScenes~\cite{caesar2020nuscenes}.} 
We evaluate the transferability of visual semantics to 3D point clouds via knowledge distillation. 
The results highlight performance gains across varying data regimes (\eg, 1\% vs. 100\% labeled data), underscoring the data efficiency of LiDAR-centric pre-training.}
    \vspace{-2mm}
    \resizebox{\textwidth}{!}{
    \begin{tabular}{r|r|c|c|p{32pt}<{\centering}p{32pt}<{\centering}p{32pt}<{\centering}p{32pt}<{\centering}p{32pt}<{\centering}p{32pt}<{\centering}|p{34pt}<{\centering}|p{34pt}<{\centering}}
    \toprule
    \multicolumn{1}{c|}{\multirow{2}{*}{\textbf{Method}}} & \multicolumn{1}{c|}{\multirow{2}{*}{\textbf{Venue}}} & \textbf{Backbone} & \textbf{Backbone} & \multicolumn{6}{c}{\textbf{nuScenes}} \vline & \textbf{KITTI} & \textbf{Waymo}
    \\
    & & \textbf{(2D)} & \textbf{(3D)} & \textbf{LP} & \textbf{1\%} & \textbf{5\%} & \textbf{10\%} & \textbf{25\%} & \textbf{Full} & \textbf{1\%} & \textbf{1\%}
    \\\midrule\midrule
    Random & - & None & MinkUNet-34 & 8.10 & 30.30 & 47.84 & 56.15 & 65.48 & 74.66 & 39.50 & 39.41
    \\\midrule
    PointContrast \cite{xie2020pointcontrast} & ECCV'20 & \multirow{4}{*}{None} & \multirow{4}{*}{\makecell{MinkUNet-34 \\ \cite{choy2019minkowski}}} & 1.90 & 32.50 & - & - & - & - & 41.10 & -
    \\
    DepthContrast \cite{zhang2021self} & ICCV'21 & & & 2.10 & 31.70 & - & - & - & - & 41.50 & -
    \\
    ALSO \cite{boulch2023also} & CVPR'23 & & & - & 37.70 & - & 59.40 & - & 72.00 & - & -
    \\
    BEVContrast \cite{sautier2024bevcontrast} & 3DV'24 & & & - & 38.30 & - & 59.60 & - & 72.30 & - & -
    \\\midrule
    PPKT \cite{liu2021learning} & arXiv'21 & \multirow{9}{*}{\makecell{~ResNet-50 \\ \cite{he2016resnet}}~} & \multirow{9}{*}{\makecell{MinkUNet-34 \\ \cite{choy2019minkowski}}} & 35.90 & 37.80 & 53.74 & 60.25 & 67.14 & 74.52 & 44.00 & 47.60
    \\
    SLidR \cite{sautier2022image} & CVPR'22 & & & 38.80 & 38.30 & 52.49 & 59.84 & 66.91 & 74.79 & 44.60 & 47.12
    \\
    ST-SLidR \cite{mahmoud2023self} & CVPR'23 & & & 40.48 & 40.75 & 54.69 & 60.75 & 67.70 & 75.14 & 44.72 & 44.93
    \\
    TriCC \cite{pang2023unsupervised} & CVPR'23 & & & 38.00 & 41.20 & 54.10 & 60.40 & 67.60 & 75.60 & 45.90 & -
    \\
    Seal \cite{liu2024segment} & NeurIPS'23 & & & 44.95 & 45.84 & 55.64 & 62.97 & 68.41 & 75.60 & 46.63 & 49.34
    \\
    CSC \cite{chen2024building} & CVPR'24 & & & 46.00 & 47.00 & 57.00 & 63.30 & 68.60 & 75.70 & 47.20 & -
    \\
    OLIVINE \cite{zhang2024fine} & NeurIPS'24 & & & 50.09 & 50.58 & 60.19 & 65.01 & 70.13 & 76.54 & 49.38 & -
    \\
    HVDistill \cite{zhang2024hvdistill} & IJCV'24 & & & 39.50 & 42.70 & 56.60 & 62.90 & 69.30 & 76.60 & 49.70 & -
    \\
    LargeAD \cite{kong2025largead} & arXiv'25 & & & 46.13 & 47.08 & 56.90 & 63.74 & 69.34 & 76.03 & 49.55 & 50.29
    \\\midrule
    PPKT \cite{liu2021learning} & arXiv'21 & \multirow{7}{*}{\makecell{ViT-S \\ \cite{oquabdinov2}}} & \multirow{7}{*}{\makecell{~MinkUNet-34 \\ \cite{choy2019minkowski}}~} & 38.60 & 40.60 & 52.06 & 59.99 & 65.76 & 73.97 & 43.25 & 47.44
    \\
    SLidR \cite{sautier2022image} & CVPR'22 & & & 44.70 & 41.16 & 53.65 & 61.47 & 66.71 & 74.20 & 44.67 & 47.57
    \\
    Seal \cite{liu2024segment} & NeurIPS'23 & & & 45.16 & 44.27 & 55.13 & 62.46 & 67.64 & 75.58 & 46.51 & 48.67
    \\
    ScaLR \cite{puy2024three} & CVPR'24 & & & 42.40 & 40.50 & - & - & - & - & - & -
    \\
    SuperFlow \cite{xu20244d} & ECCV'24 & & & 46.44 & 47.81 & 59.44 & 64.47 & 69.20 & 76.54 & 47.97 & 49.94
    \\
    LargeAD \cite{kong2025largead} & arXiv'25 & & & 46.58 & 46.78 & 57.33 & 63.85 & 68.66 & 75.75 & 50.07 & 50.83
    \\
    LiMoE \cite{xu2025limoe} & CVPR'25 & & & 48.20 & 49.60 & 60.54 & 65.65 & 71.39 & 77.27 & 49.53 & 51.42
    \\
    \bottomrule
\end{tabular}}
\label{table:nus_image2lidar}
\end{table}

\subsubsection{LiDAR Segmentation}
LiDAR semantic segmentation, involving dense point-level classification, is the rigorous testing ground for the \textbf{Semantic-Geometric Gap}. 
Since point clouds inherently lack texture, performance on this task directly reflects a model's ability to hallucinate semantics from geometry.
The comparisons in Table~\ref{table:nus_image2lidar}  reveal a decisive trend: \textbf{Camera-to-LiDAR Distillation} is indispensable, particularly for \textbf{Data Efficiency}.

Approaches utilizing visual priors consistently surpass training-from-scratch baselines, with advantages magnified in data-scarce regimes.
For instance, with only $1\%$ of labeled data, the random baseline yields a poor mIoU of $30.30$. In stark contrast, distillation-based methods like \textbf{OLIVINE}~\cite{zhang2024fine} and \textbf{LiMoE}~\cite{xu2025limoe} achieve $50.58$ and $49.60$ mIoU respectively, effectively doubling the performance of the baseline. This indicates that self-supervised pre-training effectively unlocks the latent geometric structure of unlabeled data, significantly reducing the dependency on costly manual annotations.

Crucially, the results uncover a \emph{Scaling Law Transfer} phenomenon. Advanced distillation methods like \textbf{LiMoE}~\cite{xu2025limoe} not only achieve state-of-the-art results on the full dataset ($77.27$ mIoU)  but also demonstrate that 3D backbones can inherit the rich, open-world semantics of large-scale 2D Foundation Models (utilizing ViT-S teachers). It validates the hypothesis that forging Spatial Intelligence does not require reinventing semantic understanding, but rather effectively transferring it from the vision domain to the 3D physical world.

\section{Open-World Perception and Planning}
\label{sec:open}

The ultimate goal of Spatial Intelligence is not merely to perceive closed-set categories but to generalize to the open world and make robust decisions in unseen scenarios. 
Traditional perception systems, constrained by fixed ontologies and supervised data, struggle with the long-tail unpredictability of real-world environments.
In this section, we explore how multi-modal pre-training is evolving to address these challenges. 
We first analyze the demands of \textbf{Open-World Perception} (Section~\ref{sec:open_chan}). We then discuss how \textbf{Text-Grounded Understanding} leverages Vision-Language Models (VLMs) to bridge the semantic gap and automate supervision (Section~\ref{sec:text_ground}). 
Finally, we examine the culmination of these efforts in \textbf{Unified World Representations}, where generative world models and Vision-Language-Action (VLA) architectures are redefining end-to-end planning (Section~\ref{sec:world_model}).

\subsection{Open-World Challenges}
\label{sec:open_chan}

Open-world deployment introduces complexity vectors that exceed the capacity of traditional representation learning:
\begin{itemize}
    \item \textbf{Open-Vocabulary Recognition:} Systems must identify novel objects (\eg, "overturned truck", "debris") that were never explicitly annotated during training, requiring a shift from ID-based classification to language-driven reasoning.
    \item \textbf{Domain Shifts and Anomalies:} Robustness against changing weather, lighting, and sensor degradation is critical. Models must quantify epistemic uncertainty to handle "unknown unknowns" safely.
    \item \textbf{Data Scalability:} The combinatorial explosion of corner cases makes manual annotation infeasible. Learning from vast, unlabeled, diverse data streams is the only viable path to coverage.
\end{itemize}
Addressing these challenges necessitates a paradigm shift: from learning specific tasks to learning generalizable \emph{world knowledge}.

\subsection{Text-Grounded Understanding}
\label{sec:text_ground}

Language serves as the universal interface for open-world knowledge. 
By aligning 3D sensor data with rich textual semantics, foundation models can \emph{read} the scene, unlocking zero-shot capabilities. 
This paradigm manifests in two key directions: \textbf{Auto-Labeling Data Engines} and \textbf{Open-Vocabulary Representation Learning}.

\noindent \textbf{Auto-Labeling as a Scalable Data Engine.}
The most immediate impact of foundation models is breaking the annotation bottleneck. 
Instead of relying on human labelers, recent works utilize pre-trained VLMs~\cite{radford2021clip, wang2024qwen2} to generate high-quality pseudo-labels for sensor data.
\textbf{CLIP2Scene}~\cite{chen2023clip2scene} and \textbf{OpenScene}~\cite{peng2023openscene} pioneered the distillation of 2D vision-language features into 3D point clouds, effectively automating semantic segmentation.
Advanced frameworks like \textbf{Affinity3D}~\cite{liu2024affinity3d} and \textbf{VLM2Scene}~\cite{liao2024vlm2scene} further refine this process by enforcing multi-view consistency, ensuring that the hallucinated labels are geometrically coherent for downstream supervised training.

\noindent \textbf{Text-Assisted Representation Learning.}
Beyond generating discrete labels, recent research focuses on \textbf{Self-Supervised 3D Occupancy Prediction}, treating text-aligned 2D features as continuous supervision signals.
Methods like \textbf{LangOcc}~\cite{boeder2024langocc} and \textbf{LOcc}~\cite{yu2024language} leverage knowledge distillation from diverse teacher models~\cite{ravi2024sam, oquabdinov2, bai2025qwen2} to directly guide the learning of dense volumetric semantics.
As shown in Table~\ref{table:nus_self_occ}, these self-supervised approaches now rival supervised baselines, proving that foundation model-driven supervision can replace manual effort.
Furthermore, the trend towards \textbf{3D Gaussian Splatting}~\cite{jiang2024gausstr, boeder2025gaussianflowocc, zhao2025shelfgaussian} illustrates the push for representations that are not only semantically rich but also geometrically continuous and renderable, facilitating better alignment with 2D VLMs.

\begin{table}
    \centering
    \renewcommand\tabcolsep{3pt}
    \caption{\textbf{Performance comparison for self-supervised 3D occupancy prediction on Occ3D-nuScenes~\cite{tian2024occ3d}.} This table assesses the capability of methods to learn dense volumetric representations without manual 3D labels. ``FM'' denotes the specific 2D Foundation Model utilized for pseudo-label generation or feature distillation.}
    \vspace{-2mm}
    \resizebox{\textwidth}{!}{
    \begin{tabular}{r|r|c|c|c|c|cc}
    \toprule
    \multicolumn{1}{c|}{\multirow{2}{*}{\textbf{Method}}} & \multicolumn{1}{c|}{\multirow{2}{*}{\textbf{Venue}}} & \multirow{2}{*}{\textbf{Representation}} & \multirow{2}{*}{\textbf{Foundation Model Used}} & \multirow{2}{*}{\textbf{Supervision}} & \multirow{2}{*}{\textbf{Other Supported Task}} & \multicolumn{2}{c}{\textbf{Performance}}
    \\ 
    ~ & ~ & ~ & ~ & ~ & ~  & IoU & mIoU   
    \\
    \midrule\midrule
    SimpleOcc~\cite{gan2024comprehensive} & TIV'24 & NeRF &  - & Video Sequence  & Depth Estimation & -  & $7.99$ \\
    OccNeRF~\cite{zhang2023occnerf} & {TIP'25} & NeRF & Grounding DINO~\cite{liu2024groundingdino} & Video Sequence \& FM & Depth Estimation & $22.81$ & $9.53$   
    \\
    SelfOcc~\cite{huang2024selfocc} & {CVPR'24} & BEV/TPV Feature & OpenSeeD~\cite{zhang2023openseed} & Video Sequence \& FM & Novel Depth Synthesi/Depth Estimation & $45.01$ & $9.30$   
    \\
    DistillNeRF~\cite{wang2024distillnerf} & {NeurIPS'24} & NeRF & CLIP~\cite{radford2021clip} \& DINOv2~\cite{oquabdinov2} & FM & Novel View Synthesi/Depth Estimation & $29.11$ & $8.93$   
    \\
    GaussianOcc~\cite{gan2024gaussianocc} & {ICCV'25} & Gaussians & Grounding DINO~\cite{liu2024groundingdino} & Video Sequence \& FM & Depth Estimation & - & $9.94$   
    \\
    GaussTR~\cite{huang2024selfocc} & {CVPR'25} & Gaussians & Metric3D~\cite{yin2023metric3d} \& CLIP~\cite{radford2021clip} \& SAM~\cite{kirillov2023sam} & FM & Open-Vocabulary Occupancy Prediction & $45.19$ & $11.70$   
    \\
    LangOcc~\cite{boeder2024langocc} & 3DV'25 &NeRF  & MaskCLIP~\cite{zhou2022maskclip}  &  Video Sequence \& FM & 3D Open Vocabulary Retrieval &  $51.76$  & $11.84$ \\
    VEON-L~\cite{zheng2024veon} & ECCV'24 & Occ & MiDAS~\cite{ranftl2020midas} \& SAN~\cite{xu2023san} \& CLIP~\cite{radford2021clip}  & FM \& LiDAR  & 3D Open Vocabulary Retrieval &  - &  $15.14$ \\
     TT-OccLiDAR~\cite{zhang2025tt} & arXiv'25 & Gaussians &  VGGT~\cite{wang2025vggt} \& OpenSeeD~\cite{zhang2023openseed} & Video Sequence \& FM \& LiDAR  &  Progressive Occupancy Estimation&  - & $23.60$ \\
     GaussianFlowOcc~\cite{boeder2025gaussianflowocc} & arXiv'25 & Gaussians & GroundedSAM~\cite{ren2024grounded} \& Metric3D~\cite{yin2023metric3d} & Video Sequence \& FM & Depth Estimation & $46.91$ & $17.08$ \\
     ShelfOcc~\cite{boeder2025shelfocc} & arXiv'25 & Voxel & MapAnything~\cite{keetha2025mapanything} \& GroundedSAM~\cite{ren2024grounded} & Video Sequence \& FM & - & $56.14$ & $22.87$ \\
     ShelfGaussian~\cite{zhao2025shelfgaussian} & arXiv'25 & Gaussians & DINOv2~\cite{oquabdinov2} \& Metric3D~\cite{yin2023metric3d} & Video Sequence \& FM \& LiDAR & BEV Segmentation / Trajectory Planning & $63.25$ & $19.07$ \\
     QueryOcc~\cite{lilja2025queryocc} &arXiv'25&Query&Metric3D~\cite{yin2023metric3d} \& GroundedSAM~\cite{ren2024grounded} \& DinoV3~\cite{simeoni2025dinov3} &Video Sequence \& FM \& LiDAR& Depth Estimation&55.00&21.30\\
    \bottomrule
\end{tabular}}
\label{table:nus_self_occ}
\end{table}

\subsection{Unified World Representation for Action}
\label{sec:world_model}

Perception serves as the foundation for decision-making, while the ultimate manifestation of Spatial Intelligence is \textbf{Action}. 
The field is transitioning from modular perception-planning pipelines to unified \textbf{World Models} that can simulate future states and plan end-to-end within a shared space.

\noindent \textbf{From Discriminative to Generative Planning.}
Traditional end-to-end planning often relied on explicit perception outputs (\eg, bounding boxes and vectorized maps) or decoupled feature maps. 
Recent breakthroughs, however, are driven by \textbf{Generative World Models}~\cite{kong20253d, zheng2025world4drive, yang2024driving}.
Moving beyond discrete label prediction, models like \textbf{OccWorld}~\cite{zheng2023occworld} and \textbf{GenAD}~\cite{yang2024generalized} learn to predict the future evolution of the 3D world (\eg, 4D Occupancy flow) conditioned on ego-actions.
This \emph{predictive learning} objective forces the model to internalize scene dynamics, causal relationships, and object interactions. 
 As evidenced in Table~\ref{table:nus_e2e}, these generative planners significantly outperform discriminative baselines in both collision rates and open-loop planning metrics.

\begin{table}
    \renewcommand\tabcolsep{3pt}
    \centering
    \footnotesize
    \caption{\textbf{Evaluation of end-to-end planning on the nuScenes benchmark~\cite{caesar2020nuscenes}.}
    The table compares the planning fidelity of state-of-the-art methods, contrasting traditional pipelines with emerging Generative World Models. 
    Performance is measured by \textbf{planning L2 error (L2)} and \textbf{Collision Rate (CR)}, where \textit{lower values indicate better} safety and precision.}
    \vspace{-2mm}
    \resizebox{\textwidth}{!}{
    \begin{tabular}{r|r|c|l|c|c|ccc}
    \toprule
    \multicolumn{1}{c|}{\multirow{2}{*}{\textbf{Method}}} & \multicolumn{1}{c|}{\multirow{2}{*}{\textbf{Venue}}} & \multirow{2}{*}{\textbf{Input}} & \multicolumn{1}{c|}{\multirow{2}{*}{\textbf{Representation}}} & \multirow{2}{*}{\textbf{Supported Task}} & \multirow{2}{*}{\textbf{Auxiliary Supervision}} & \multicolumn{3}{c}{\textbf{Performance}} 
    \\ 
    ~ & ~ & ~ & ~ & ~ & ~  & L2 Avg. (m) & CR Avg. & FPS  
    \\
    \midrule\midrule
    ST-P3 \cite{hu2022st} & {ECCV'22} & Image & \textcolor{cameragreen}{$\bullet$}~BEV Feature & BEV Seg. & Map \& Box \& Depth & 2.11 & 0.71 & 1.6
    \\
    UniAD \cite{hu2023planning} & {CVPR'23} & Image & \textcolor{cameragreen}{$\bullet$}~BEV Feature &  Track./Map/Motion Fore./Occ. &  Map \& Box \& Motion \& Tracklets \& Occ & 1.03 & 0.31 & 1.8  
    \\
    VAD-Tiny \cite{jiang2023vad} & {ICCV'23} & Image & \textcolor{cameragreen}{$\bullet$}~Vectorized BEV Scene & Vectorized Map & Map \& Box \& Motion & 1.30 & 0.72 & 16.8
    \\
    VAD-Base \cite{jiang2023vad} & {ICCV'23} & Image & \textcolor{cameragreen}{$\bullet$}~Vectorized BEV Scene & Vectorized Map & Map \& Box \& Motion & 1.22 & 0.53 & 4.5  
    \\
    OccNet \cite{tong2023scene} & {ICCV'23} & Image & \textcolor{lidarblue}{$\bullet$}~3D Occupancy & Semantic Occ. Pred. & 3D-Occ \& Map \& Box & 2.14 & 0.72 & 2.6  
    \\
    \midrule
    OccWorld \cite{zheng2023occworld} & {ECCV'24} & Image & \textcolor{lidarblue}{$\bullet$}~3D Occupancy & 4D Occ. Fore. & 3D-Occ & 1.34 & 0.73 & 2.8  
    \\
    OccWorld \cite{zheng2023occworld} & {ECCV'24} & Image & \textcolor{lidarblue}{$\bullet$}~3D Occupancy & 4D Occ. Fore. & None & 1.83 & 2.02 & 2.8
    \\
    OccWorld\cite{zheng2023occworld} & {ECCV'24} & Occ & \textcolor{lidarblue}{$\bullet$}~3D Occupancy & 4D Occ. Fore. & None & 1.17 & 0.60 & 18.0
    \\
    RenderWorld \cite{yan2024renderworld} & {ICRA'25} & Image & \textcolor{lidarblue}{$\bullet$}~3D Occupancy & 4D Occ. Fore. & None & 1.48 & 0.97 & -  
    \\
    RenderWorld \cite{yan2024renderworld} & {ICRA'25} & Occ & \textcolor{lidarblue}{$\bullet$}~3D Occupancy & 4D Occ. Fore. & None & 1.03 & 0.61 & -  
    \\
    OccLLaMA \cite{wei2024occllama} & {arXiv'24} & Image & \textcolor{lidarblue}{$\bullet$}~3D Occupancy & 4D Occ. Fore./VQA & 3D-Occ & 1.20 & 0.70 & -  
    \\
    OccLLaMA \cite{wei2024occllama} & {arXiv'24} & Occ & \textcolor{lidarblue}{$\bullet$}~3D Occupancy & 4D Occ. Fore./VQA & None & 1.14 & 0.49 & -  
    \\
    OccVAR \cite{jinoccvar} & {arXiv'24} & Image & \textcolor{lidarblue}{$\bullet$}~3D Occupancy & 4D Occ. Fore. & 3D-Occ & 1.35 & 0.83 & -  
    \\
    OccVAR \cite{jinoccvar} & {arXiv'24} & Occ & \textcolor{lidarblue}{$\bullet$}~3D Occupancy & 4D Occ. Fore. & None & 1.21 & 0.78 & -  
    \\
    LAW \cite{li2024enhancing} & {ICLR'25} & Image & \textcolor{cameragreen}{$\bullet$}~Latent Feature & Latent Prediction Fore. & None & 0.61 & 0.30 & 19.5  
    \\
    SSR \cite{li2024does} & {ICLR'25} & Image & \textcolor{cameragreen}{$\bullet$}~BEV Feature & BEV Feature Fore. & None & 0.39 & 0.06 & 19.5  
    \\
    FSF-Net \cite{guo2024fsf} & {arXiv'24} & Occ & \textcolor{lidarblue}{$\bullet$}~3D Occupancy & 4D Occ. Fore. & None & 0.82 & 0.01 & -  
    \\
    Drive-OccWorld \cite{yang2024driving} & {AAAI'25} & Image & \textcolor{lidarblue}{$\bullet$}~3D Occupancy & 4D Occ. Fore./Generation & 3D-Occ & 0.85 & 0.29 & -  
    \\
    OccTens \cite{jin2025occtens} & {arXiv'25} & Occ & \textcolor{lidarblue}{$\bullet$}~3D Occupancy & 4D Occ. Fore./Generation & 3D-Occ & 1.12 & 0.48 & -  
    \\
    OccVLA \cite{liu2025occvla} & {arXiv'25} & Occ & \textcolor{lidarblue}{$\bullet$}~3D Occupancy & 3D Occ. Generation & 3D-Occ & 0.28 & - & -  \\
     World4Drive \cite{zheng2025world4drive} & {ICCV'25} & Image & \textcolor{cameragreen}{$\bullet$}~Latent Feature & Latent Prediction Fore. & Open-vocabulary Semantics & 0.50 & 0.16 & -  
    \\
    \bottomrule
\end{tabular}}
\label{table:nus_e2e}
\end{table}

\noindent \textbf{Unified End-to-End Architectures: VA and VLA.}
The convergence of generative modeling and autonomous driving has bifurcated into two powerful paradigms for action generation: \textbf{Vision-Action (VA)} latent models and \textbf{Vision-Language-Action (VLA)} reasoning frameworks.
The first paradigm focuses on pure decision-making efficiency by constructing \textbf{Latent World Models}. 
Unlike traditional pipelines that rely on explicit perception supervision, methods like \textbf{LAW}~\cite{li2024enhancing} and \textbf{SSR}~\cite{li2024does} bypass human annotations entirely. 
By abstracting the environment into high-dimensional latent states, these models learn to predict future rewards and control signals directly from sensor inputs without the need for perception labels.

Parallel to pure latent modeling, the integration of Large Language Models (LLMs) has catalyzed the emergence of \textbf{VLA} frameworks that emphasize interpretability and open-world reasoning~\cite{liu2023llava, bai2025qwen2, wang2025pixelthink}.
Approaches like \textbf{OccVLA}~\cite{liu2025occvla} and \textbf{DriveVLA-W0}~\cite{li2025drivevla} tokenize visual input and project them into the LLM's context window alongside text.
This enables the system to not only generate control actions but also to perform causal reasoning (``\emph{Why is the car stopping?}'') and handle complex social interactions (``\emph{Yield to the aggressive merger}'') in a unified autoregressive process.

In summary, the trajectory is clear: from \emph{detecting objects} to \emph{simulating latent futures (VA)}, and finally to \emph{reasoning with language (VLA)}. 
This evolution underscores the pivotal role of multi-modal pre-training in constructing the next generation of embodied intelligent systems.

\section{Challenges and Future Directions}
\label{sec:chall_and_future}
As demonstrated in this work, the pursuit of Spatial Intelligence has evolved from task-specific supervision to a paradigm dominated by large-scale, multi-modal pre-training. 
While the techniques analyzed in Section~\ref{sec:methods} and Section~\ref{sec:open} demonstrate immense progress, the rapid emergence of generative AI and foundation models introduces new frontiers. 
In this section, we synthesize critical remaining obstacles and outline a forward-looking research agenda centered on generative world modeling and embodied reasoning.

\subsection{Current Challenges}

\noindent \textbf{The Semantic-Geometric Gap.} 
A fundamental dissonance remains between the rich semantic knowledge encapsulated in Vision-Language Models (VLMs) and the precise metric requirements of autonomous control. 
While VLMs excel at open-vocabulary recognition~\cite{radford2021clip, wang2024qwen2, bai2025qwen2, liu2023llava}, they often lack the fine-grained spatial grounding necessary to localize it with centimeter-level accuracy. 
Bridging the gap between high-level semantic reasoning and low-level geometric constraints without compromising either remains a formidable theoretical and engineering challenge~\cite{liu2025occvla, wei2024occllama}.

\noindent \textbf{Data-Centric Bottlenecks and Corner Cases.} 
The scaling laws of foundation models are increasingly hitting diminishing returns regarding data quality. 
The primary challenge has shifted from acquiring \emph{more} data to mining \emph{valuable} data—specifically, long-tail corner cases and safety-critical scenarios~\cite{ghosh2024roadwork, yurt2025ltda, tian2024tokenize}. 
Current pre-training objectives treat all data samples equally, often wasting computation on repetitive driving patterns while under-weighting rare, high-value events~\cite{yang2023vidar,yang2024unipad}. 
Furthermore, utilizing foundation models for auto-labeling introduces epistemic uncertainty that is difficult to filter from the training pipeline.

\noindent \textbf{Real-Time Inference of Foundation Models.} 
There is a growing disparity between the computational demands of state-of-the-art pre-trained models and the strict latency/power constraints of onboard edge devices~\cite{jihong2024edge, weng2024drive}. 
While cloud-based pre-training leverages unlimited resources, distilling these massive \emph{teacher} models into lightweight, real-time \emph{student} networks without catastrophic performance drops is an ongoing bottleneck for deployment~\cite{zhang2024hvdistill, xu20244d, xu2025beyond}.

\subsection{Future Directions}
\label{sec:future_directions}

While recent advancements have laid the foundation for spatial intelligence, several critical frontiers remain to be conquered to achieve robust, human-level autonomy.

\noindent \textbf{Physically Consistent World Simulators.} 
Although emerging Generative World Models~\cite{kong20253d, zheng2025world4drive, li2025drivevla} can synthesize plausible futures, they often suffer from hallucinations that violate physical laws~\cite{worldlens}. 
A key future direction is to enforce \emph{Physical Consistency} within the pre-training objective. 
By integrating differentiable physics engines or explicit geometric constraints into the generation process~\cite{li2025drivevla, wang2024drivedreamer, bartoccioni2025vavim, liu2025occvla}, future models must evolve from merely generating visual pixels to simulating realistic physical interactions, thereby serving as reliable training environments for safety-critical policies.

\noindent \textbf{Trustworthy and Real-Time Embodied VLA.} 
Current Vision-Language-Action (VLA) models~\cite{li2025drivevla, jiang2025survey, liu2025occvla, survey_vla4ad} demonstrate promise but face significant hurdles in real-world deployment: high inference latency and lack of interpretability. 
Future research should bridge the gap between heavy foundation models and the millisecond-level reaction requirements of autonomous systems. 
This necessitates exploring lightweight VLA architectures, efficient tokenization strategies, and mechanisms for uncertainty quantification to ensure that end-to-end decision-making is not only intelligent but also trustworthy and verifiable~\cite{wang2024reliocc, yu2025inst3d, li2025tokenpacker, feng2025can}.

\noindent \textbf{4D Semantic-Geometric Unification.} 
The transition from discrete voxels to continuous representations like 3D Gaussian Splatting (3DGS)~\cite{kerbl20233d, jiang2024gausstr, liu2025gaussian2scene} is underway. 
However, current 3DGS methods largely focus on visual rendering quality rather than semantic understanding. 
The next frontier lies in \emph{Semantic Lifting}—imbuing these continuous geometric primitives with dense semantic and instance-level attributes over time. 
Pre-training tasks that enforce spatiotemporal consistency on Gaussian attributes~\cite{xu2024gaussianpretrain, boeder2025gaussianflowocc} will be pivotal for enabling agents to not just view the scene, but to manipulate and interact with specific objects in a dynamic 4D world.

\noindent \textbf{System 2 Reasoning for Long-Tail Safety.} 
Existing pre-training paradigms excel at pattern recognition (\emph{System 1}) but struggle with rare, complex scenarios requiring logical deduction. 
Future systems will integrate \emph{System 2} capabilities~\cite{tian2024drivevlm, survey_vla4ad}, potentially via Chain-of-Thought (CoT) distillation from LLMs~\cite{wei2022chain, zhai2025world, wang2025omnidrive, wang2025pixelthink}. 
The goal is to move beyond passive explanation to active \emph{Causal Reasoning}—enabling the vehicle to counterfactually simulate \emph{what if} scenarios and override reactive policies when facing novel, long-tail safety hazards.

\section{Conclusion}
\label{sec:conc}

In this paper, we have presented a systematic analysis of multi-modal pre-training for autonomous systems, characterizing the evolution from modality-specific pre-training to unified foundation models as the cornerstone of \emph{Spatial Intelligence}. 
By structuring datasets and methodologies across autonomous vehicles, drones, and other robotic systems, we demonstrated how integrating complementary sensor modalities (specifically camera and LiDAR) creates representations that are both semantically rich and geometrically precise. 
Our analysis confirms that leveraging pre-trained foundation models is no longer optional but essential for achieving open-world generalization and mitigating the scarcity of annotated 3D data.

Looking ahead, the field stands at a critical inflection point. 
As demonstrated, the paradigm is shifting from passive perception to active, embodied reasoning. 
Future breakthroughs will likely stem from bridging the semantic-geometric gap through \textbf{Generative World Models} that serve as neural simulators, and from the development of end-to-end \textbf{Vision-Language-Action (VLA)} frameworks that unify perception with decision-making. Furthermore, equipping these systems with explicit reasoning capabilities will be pivotal for handling the long-tail unpredictability of real-world environments.
Ultimately, the transition from \emph{seeing} to \emph{acting} and \emph{reasoning} represents the next frontier. Continued advancements in these generative and embodied pre-training paradigms will be instrumental in forging autonomous systems that are not only robust and scalable but possess true Spatial Intelligence for safe and real-world deployment.

\bibliographystyle{plainnat}
\bibliography{reference}

\end{document}